\documentclass[runningheads]{llncs}
\usepackage[T1]{fontenc}
\usepackage{graphicx}
\usepackage{booktabs}
\usepackage[misc]{ifsym}

\usepackage{mwe}

\usepackage{algorithm}
\usepackage{algorithmic}
\usepackage{amsmath}
\usepackage{amssymb}
\usepackage{mathtools}
\usepackage{paracol}
\usepackage{multirow}
\usepackage{tcolorbox, mathtools}
\tcbuselibrary{skins}
\usepackage{enumitem}

\newtcolorbox{mybox}[1][]{before=\centering, drop fuzzy shadow, enhanced, colback=white, sharp corners, colframe=red, fonttitle=\bfseries, title=#1, center title}

\usepackage{pifont}
\newcommand{\xmark}{\ding{55}}%

\usepackage{tikz}
\usetikzlibrary{decorations.pathreplacing,calc}
\newcommand{\tikzmark}[1]{\tikz[overlay,remember picture] \node (#1) {};}

\newcommand*{\AddNote}[4]{%
    \begin{tikzpicture}[overlay, remember picture]
        \draw [decoration={brace,amplitude=0.5em},decorate,ultra thick]
            ($(#3)!(#1.north)!($(#3)-(0,1)$)$) --  
            ($(#3)!(#2.south)!($(#3)-(0,1)$)$)
                node [align=center, text width=2.5cm, pos=0.5, anchor=west] {#4};
    \end{tikzpicture}
}%
\usepackage{hyperref}
\hypersetup{
    colorlinks,
    linkcolor={green!70!black},
    citecolor={blue!70!black},
    urlcolor=magenta
}

\usepackage[colorinlistoftodos,prependcaption,textsize=footnotesize]{todonotes}

\newcommand{\Rho}{\mathrm{P}}

\begin{document}

\title{Producer-Fairness in Sequential Bundle Recommendation}

\titlerunning{Producer-Fairness in Sequential Bundle Recommendation}

\author{Alexandre Rio\inst{1,3}
\and Marta Soare\inst{2}
\and Sihem Amer-Yahia\inst{3}}


\institute{Huawei Noah's Ark Lab, Paris, France
\and Université d'Orléans, France
\and Université Grenoble-Alpes, France}

\maketitle              

\begin{abstract}
We address fairness in the context of sequential bundle recommendation, where users are served in turn with sets of relevant and compatible items. Motivated by real-world scenarios, we formalize producer-fairness, that seeks to achieve desired exposure of different item groups across users in a recommendation session. Our formulation combines naturally with building high quality bundles.
Our problem is solved in real time as users arrive. We propose an exact solution that caters to small instances of our problem. We then examine two heuristics, quality-first and fairness-first, and an adaptive variant that determines on-the-fly the right balance between bundle fairness and quality. Our experiments on three real-world datasets underscore the strengths and limitations of each solution and demonstrate their efficacy in providing fair bundle recommendations without compromising bundle quality. 

\end{abstract}

\section{Introduction}
\label{sec:intro}
Today, an increasing number of marketplaces serve content in the form of bundles, \textit{i.e.}, a composite set of items that are not only relevant to the user (or consumer), but also compatible with each other (\textit{e.g.}, points of interest within walking distance),
and complementary (\textit{e.g.}, movies of different genre). Bundle examples are plentiful and include books on the same topics, souvenir baskets, and travel packages~\cite{Amer-YahiaBCFMZ14,BenouaretL16,Chang23,DBLP:conf/sigmod/RoyACDY10,XieLW12}. 
In this work, we are interested in marketplaces where one bundle is served to one user at a time in a sequence~\cite{MehrotraMBL018}. 

While serving high-quality bundles to  users is often the primary goal, ensuring fair visibility of different items is also essential in recommendation systems. Fairness in single-item top-$k$ recommendations is most commonly defined as fairness of exposure of items in a ranked list~\cite{DBLP:journals/vldb/PitouraSK22,SinghJ18}. 
In our work, we focus on P(roducer)-fairness in bundles, aiming to achieve fair exposure of items from different producers. While users arrive sequentially and seek the highest-quality bundle, producers expect fair representation of their products over time --- resulting in potentially conflicting objectives.
In this context, formalizing and enforcing  fair exposure of items from different groups is a challenge.

\begin{figure}[t]
\centering
\includegraphics[width=0.9\linewidth]{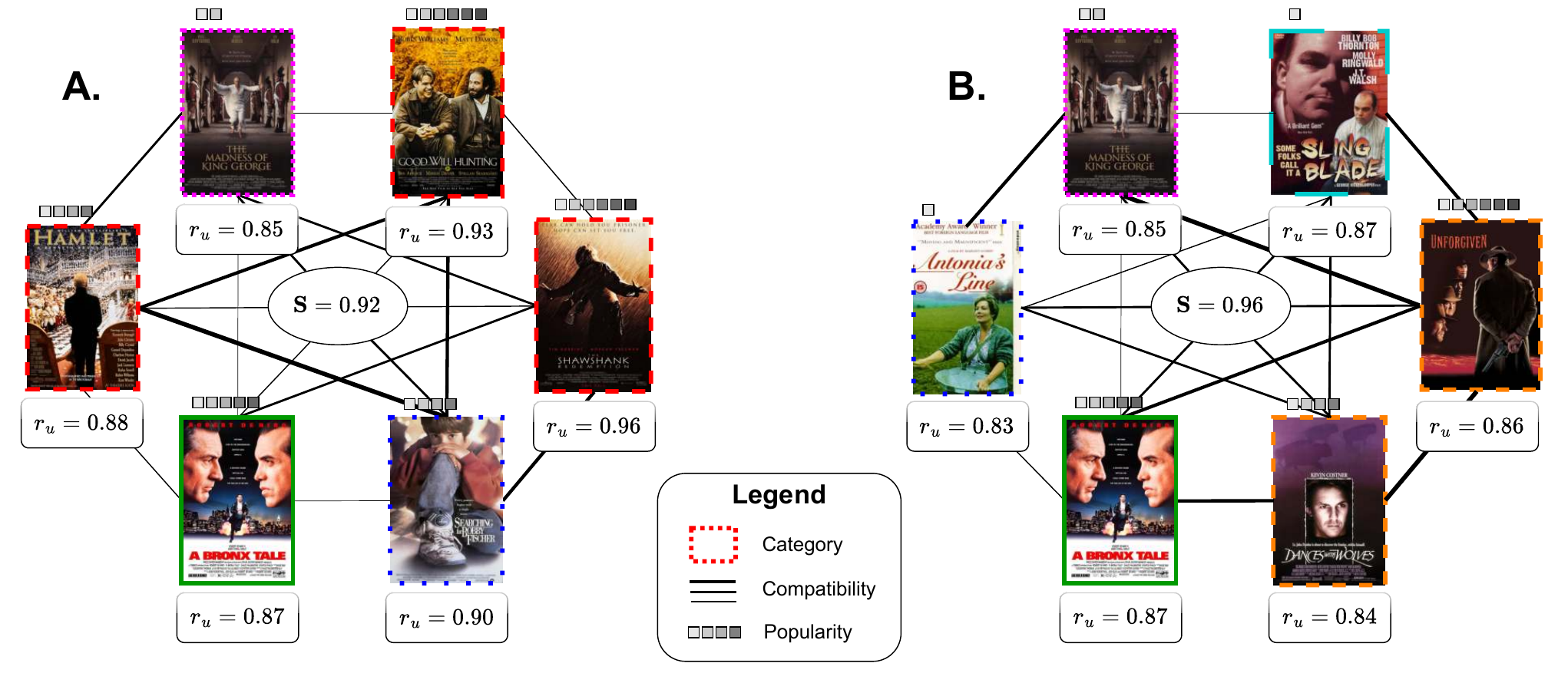}

\caption{Two examples of movie bundles (from \textit{MovieLens}). Here, fairness is defined with respect to movie popularity. The most popular movies may be unfairly over-exposed as a result of the well-known popularity bias of recommendation systems.}
\label{fig:bundle_rec_schema}

\end{figure}

\noindent{\bf Challenges.} 
Our first challenge {\bf C1} is to devise a problem formulation that 1) combines bundle quality and P-fairness, 2) is expressive enough to cover a wide range of scenarios, and 3) can be effectively tackled. 
This is not straightforward because we must optimize various dimensions of bundle quality for each user --—relevance, compatibility, and complementarity—-- while also accounting for fairness, which is defined over the entire recommendation sequence.
Figure~\ref{fig:bundle_rec_schema} illustrates the different dimensions of the problem through two examples of movie bundles, highlighting its complexity. 
In particular, fairness methods designed for single-item recommendation are not suitable: even if a bundle is built from items produced by a fair top-$k$ algorithm, accounting for compatibility and complementarity constraints may lead to choosing items that do not achieve the desired exposure.
Our second challenge {\bf C2} is computational. As shown in the literature, building the highest quality bundle is expensive in itself and often NP-hard~\cite{Amer-YahiaBCFMZ14,DBLP:conf/sigmod/RoyACDY10}, as we must consider pairwise compatibility between items. 
Adding fairness satisfaction --- especially as it is defined across multiple users --- further increases its complexity.

\noindent{\bf Contributions.} 
To address {\bf C1}, we organize items into groups and adapt the notion of demographic parity of exposure~\cite{SinghJ18}. To achieve that, we define an exposure vector that reflects the desired exposure of each group. This allows us to formalize the P-Fair Bundle Recommendation Problem which seeks to build the highest quality bundle for each user such that the sequence of recommended bundles satisfies the exposure vector. This formulation is highly expressive as it covers common cases such as paid inclusion, that is, when a producer pays for a number of listings of their products, or a legal requirement that imposes exposing certain item categories.  To address {\bf C2}, we first propose \textsc{ILP}, an exact Integer Linear Programming solution.  This formulation is NP-hard as it generalizes the Quadratic Knapsack Problem \cite{Gallo1980} with multiple weight constraints and a cardinality constraint. 
We therefore explore two heuristic-based solutions: \textsc{F3R}, a \textit{fairness-first} approach, and \textsc{FairWG}, a \textit{quality-first} approach. 
\textsc{F3R} constructs bundles iteratively by first sampling a group of items from the exposure vector distribution, and then greedily choosing an item from this group that best improves bundle quality. \textsc{FairWG} first includes the most relevant item in the bundle and then it uses a linear combination of quality and fairness to complete it. This combination encourages the selection of items from under-exposed groups and reduces the likelihood of choosing items from over-exposed groups. \textsc{FairWG} requires to set a fixed weight to balance bundle quality and fairness. We hence introduce \textsc{AdaFairWG} that dynamically adapts the weight based on how well the current exposures align with the target exposure vector. Given a 
tolerance level, \textsc{AdaFairWG} chooses to increase or decrease the importance of fairness according to what has been achieved so far in a session.

We conduct extensive experiments on three datasets: \textit{MovieLens}, \textit{Yelp}, and \textit{Amazon}. Our results investigate the impact of P-fairness on bundle quality and explore response time and scalability. We show that our solutions enhance bundle fairness over fairness-agnostic baselines. As expected, introducing P-fairness adversely impacts quality, revealing the need to achieve a trade-off between constructing bundles and enforcing P-fairness.
\textsc{ILP} generally achieves the best quality-fairness trade-off but is only applicable to small instances in practice.
We observe that \textsc{FairWG} outperforms \textsc{F3R} and can actually achieve better trade-offs than \textsc{ILP} for comparable response times.
Finally, dynamically adapting the fairness weight in \textsc{AdaFairWG} provides the means to target specific fairness levels over long sessions while achieving high bundle quality.

\section{Bundle and Producer-Fairness}
\label{sec:setting}
We consider a marketplace offering \textbf{items} $\mathcal{I}$. Each item $i \in \mathcal{I}$ has a \textbf{type} $z \in \{1,...,Z\}$ (corresponding to values of an attribute or a set of attributes). We set $a_{iz} = 1$ if item $i$ is of type $z$, and $0$ otherwise. Items can belong to more than one type. 
Moreover, we consider that items come from one of $K$ groups of producers $\mathcal{G} = \{G_1,...,G_K\}$, and we denote $g_{ik} = \mathbf{1}(i \in G_k)$, where $\mathbf{1}(\cdot)$ is the indicator function, as well as $k(i)$ the index of item $i$'s group (\textit{i.e.}, $g_{ik(i)} = 1$). 

\subsection{Bundle} \label{sec:bundle}

Informally, a bundle $B \subset \mathcal{I}$ is a subset of items that together are relevant to the user, compatible with each other, and valid, i.e., satisfy some application-dependent constraints. 
We consider any single-item recommender which provides the top-$M$ items $\text{top}(M, u) = (i_{u(1)}, ..., i_{u(M)})$ (with $M$ large enough) as well as their relevance scores $\{r_{ui} \vert i \in \text{top}(M, u)\}$. We denote $s: \mathcal{I} \times \mathcal{I} \longrightarrow [0,1]$ a compatibility function over pairs of items. The closer $s_{ij} := s(i,j)$ is to 1, the more compatible $i$ and $j$ are.

\begin{definition} \label{def:bundle-quality} \textbf{Bundle Quality.}
The quality $\mathbf{Q}(B_u)$ of a bundle $B_u$ served to a user $u$ is a combination of aggregated (user-dependent) relevance $\mathbf{R}(B_u) = f\left(\{r_{ui} \vert i \in B_u\}\right)$, and aggregated (user-independent) pairwise item compatibilities $\mathbf{S}(B_u) = g \left(\{s_{ij} \vert i, j \in B_u\}\right)$, where $f$ and $g$ are aggregation functions.
\end{definition}

Our notion of compatibility covers many practical cases. Consider $d$ a distance $d(i,j) \in [0,1]$ between pairs of items. If we want bundles of highly similar items, we can use $s_{ij} = 1 - d(i,j)$. In contrast, we can use $s_{ij} = d(i,j)$ if we want bundles of diverse items. $s_{ij}$ can also represent a semantic notion of compatibility such as between a smartphone and a charger.

\begin{definition} \label{def:bundle-validity} \textbf{Bundle Validity.}
A bundle $B_u$ is valid if it satisfies: 
\begin{itemize}[noitemsep,topsep=0pt,leftmargin=0.15in]
    \item \!\!\textbf{Size:} its size is exactly $L$: $\vert B_u \vert = L$.
    \item \!\!\textbf{Complementarity:} it contains at most $L_z$ \!items of type $z$: $\sum_{i \in B_u} a_{iz} \le L_z$.
\end{itemize}
\end{definition}

Adopting bundles of fixed size is a pragmatic choice, as platforms typically present a predetermined number of items simultaneously.
The complementarity constraint applies to many scenarios where diversity and minimal redundancy are essential, and also particularly when bundles center on a primary item with complementary accessories --— such as a smartphone paired with a compatible case, charger, or headphone.

\subsection{Exposure and P-Fairness}
We consider a recommendation session with a horizon $T$, denoted $\mathcal{R}_T$, where we serve users $u = 1, ..., T$ sequentially (\textit{i.e.}, one by one). Each user $u$ receives a bundle $B_u$. We want to achieve fairness of exposure across different groups of producers $G_1, ..., G_K$ over time (\textit{i.e.}, across many users). Our focus here is to formalize this notion of fairness when serving bundles \textbf{in a sequence}. 

\begin{definition} \label{def:exposure} \textbf{Group Exposure and Exposure Vector.} Let $N = \sum_{u=1}^T \vert B_u \vert$ denote the total number of items recommended during a session $\mathcal{R}_T$, and $N_k = \sum_{u=1}^T \sum_{i\in B_u} g_{ik}$ denote the total number of items recommended from group $G_k$.
The exposure $E_k = \frac{N_k}{N}$ of a group $G_k$ is defined as the proportion of recommended items from $G_k$ among all items recommended during session $\mathcal{R}_T$.
An exposure vector $\Rho = \langle\rho_1, \ldots, \rho_K \rangle$, where $\rho_k \in [0, 1]$ such that $\sum_{k=1}^K \rho_k = 1$, designates the desired exposure of a set of groups $\cal{G}$ where each $\rho_k$ represents the target exposure of group $G_k$ in a recommendation session $\mathcal{R}_T$.
\end{definition}

Exposure vectors capture many cases. For instance, a producer could pay for a given number of listings of their products (\textit{i.e.}, \textit{promoted listings}), and $\rho_k$ would therefore represent the share of listings promoted by producers from group $k$. The targets could also be defined based on a normative idea or a legal requirement about product exposure: for instance, a platform could be willing to promote more items from less popular producers or might be required to expose a certain proportion of domestic products. $\rho_k$ could also simply reflect the number of products offered by $G_k$.

\begin{definition} \label{def:demographic_parity} \textbf{P(Producer)-Fairness.} Given an exposure vector $\Rho \!=\! \langle\rho_1, \ldots, \rho_K \rangle$, a recommendation session $\mathcal{R}_T$ is said to achieve perfect P-fairness iff:
    \[
        \frac{N_k}{\rho_k} = \frac{N_\ell}{\rho_\ell} \Longleftrightarrow \frac{E_k}{\rho_k} = \frac{E_\ell}{\rho_\ell} \enspace, \quad \text{ where } k,\ell = 1,...,K \enspace .
    \]
\end{definition}

Our P-fairness adapts the standard fairness notion of \textit{demographic parity} with respect to an exposure vector $\Rho$~\cite{SinghJ18}. This is equivalent to setting $E_k = \rho_k$ for all $k=1,..,K$, which is itself equivalent to $E_k \ge \rho_k$ for all $k=1,..,K$ (proof is provided in the Appendix, Section~\ref{sec:proofs}). Therefore, a recommendation session is perfectly fair if the realized exposure of any group $G_k$ is equal to its target exposure. However, achieving perfect fairness is unrealistic. We hence introduce a tolerance level $\epsilon$, such that we achieve P-fairness iff:
\[
    E_k \ge \rho_k (1 - \epsilon) \enspace, k=1,...,K \enspace
\]
where $\epsilon$ could, for instance, be chosen based on legal regulations. 


\section{The P-Fair Bundle Problem}
\label{sec:prbms}




\subsection{The P-fair Bundle Recommendation Problem} \label{sec:global_problem}

In many recommendation applications, users arrive sequentially and must be served in real time. Given a time horizon $T$ and a user $u$, the served bundle $B_u$ must be of highest relevance and compatibility while being valid: 
\begin{align}
    &\max_{B_u} \mathbf{R}(B_u) \quad \textit{(Relevance)} \label{eq:relevance}\\
    &\max_{B_u} \mathbf{S}(B_u) \quad  \textit{(Compatibility)} \label{eq:compatibility}\\
    \text{s.t. } &B_u \text{ is valid} \quad \textit{(Validity)} \label{eq:validity} \enspace.
\end{align}

Moreover, given an exposure vector $\Rho=(\rho_1, ..., \rho_K)$, we must ensure the following P-fairness constraint over the entire recommendation session $\mathcal{R}_T$:
\begin{align}
    &E_k \ge \rho_k(1 - \epsilon) \enspace, \quad \forall k=1,...,K \quad \textit{(P-fairness)} \enspace. \label{eq:const_global_pfair} 
\end{align}

While (\ref{eq:relevance})-(\ref{eq:validity}) only depend on  $u$, (\ref{eq:const_global_pfair}) is defined across all users, and cannot be addressed in its present form since ensuring (\ref{eq:const_global_pfair}) would require prior knowledge of all users in the session and the ability to compute their recommendations simultaneously.
Moreover, the conflicting nature of the two objectives (\ref{eq:relevance}) and (\ref{eq:compatibility}) adds considerable complexity to solving the problem. To address that, we propose a relaxed version of the problem that aligns with real-world circumstances.


\subsection{A Relaxed Formulation} \label{sec:cond_problem}

We propose to combine bundle relevance and compatibility into a scalarized \textit{quality} score, transforming the task into a single-objective optimization problem:
\begin{align*}
        \mathbf{Q}(B_u) &= (1 - \gamma) \times \mathbf{R}(B_u) + \gamma \times \mathbf{S}(B_u) \\
        &= (1 - \gamma) \times \frac{1}{L} \sum_{i \in B_u} r_{ui} + \gamma \times \frac{2}{L \left(L - 1 \right)} \sum_{\substack{i,j \in B_u \\ i<j}}s_{ij} ,
    \end{align*}
where $\gamma \in [0,1]$ is used to balance compatibility and relevance. 

We also transform the P-fairness constraint (\ref{eq:const_global_pfair}) into an online stepwise conditional constraint. We introduce the following quantities:
\begin{align*}
\Delta_k(u) &= \sum_{i \in B_u} g_{ik} - \vert B_u \vert \rho_k(1 - \epsilon) \enspace, \\ \Delta_k(1\!:\!u-1) &= \sum_{t=1}^{u-1}\sum_{i \in B_t} g_{ik} - \sum_{t=1}^{u-1}\vert B_t\vert\rho_k(1-\epsilon) \enspace.
\end{align*}
$\Delta_k(u)$ and $\Delta_k(1\!:\!u-1)$ are the gaps between the actual and the required number of recommended items from $G_k$ if we wanted to strictly meet the target exposure $\rho_k(1 - \epsilon)$, respectively for user $u$ (current recommendation) and for users $1$ to $u-1$ (past recommendations).
We show that if the constraint:
\begin{align}
    &\Delta_k(u) + \Delta_k(1\!:\!u-1) \ge 0 \enspace, \quad \forall k=1,...,K\enspace,\label{eq:const_cond_pfair}
\end{align}
is met for every user $u=1,...,T$, then the constraint (\ref{eq:const_global_pfair}) holds. Indeed, since $\Delta_k(u) + \Delta_k(1\!:\!u-1) = \Delta_k(1\!:\!u)$, (\ref{eq:const_cond_pfair}) is equivalent to $E_k(1\!:\! u) \ge \rho_k(1- \epsilon)$, where $E_k(1\!:\! u)$ is the exposure of $G_k$ up to user $u$. As it holds for every $u$, in particular $E_k(1\!:\!T) = E_k \ge \rho_k(1 - \epsilon)$.
\footnote{We can further interpret (\ref{eq:const_cond_pfair}) as follows: the more the group $G_k$ has been over-exposed up to the current round of recommendation (\textit{i.e.}, large $\Delta_k(1\!:\!u-1) \ge 0$), the less items from $G_k$ have to be included into $u$'s bundle to meet the P-fairness constraint, relatively to the target exposure $\rho_k(1 - \epsilon)$. Therefore, this constraint leaves room to prioritize bundle relevance and compatibility over fairness when possible.} 
We provide more details regarding how (\ref{eq:const_cond_pfair}) is derived from (\ref{eq:const_global_pfair}) in the Appendix (Section~\ref{sec:proofs}).

We replace (\ref{eq:const_global_pfair}) by the conditional constraint (\ref{eq:const_cond_pfair}) which can be computed at every round since it only depends on the current user's information and the recommendations of previous users.
We thus obtain an online optimization problem:

\begin{tcolorbox}[title = The P-fair Bundle Recommendation Problem]
\abovedisplayskip=0pt
\belowdisplayskip=0pt
For every user $u=1,...,T$, solve:
    \begin{align} \label{eq:online_opt_pbm}
&\max_{B_u} \mathbf{Q}(B_u)   \nonumber \\
\text{s.t. } &B_u \text{ is valid},  \nonumber \\
&\Delta_k(u) + \Delta_k(1\!:\!u-1) \ge 0 \enspace, \quad \forall k=1,...,K\enspace.
    \end{align}
\end{tcolorbox}

\paragraph{Time-dependent tolerance.} Since our goal is to achieve P-fairness over the entire recommendation session, it is not needed that $E_k(1\!:\!u) \ge \rho_k(1 - \epsilon)$ for $u < T$. It may in fact be suboptimal to impose such a condition, as recommendations served earlier in a session $R_T$ may benefit from more emphasis on bundle quality at the expense of fairness. To allow this while still meeting the P-fairness requirements at horizon $T$, we propose a time-dependent, monotonically decreasing tolerance $\epsilon(t)$ such that $\epsilon(t) \ge \epsilon(t+1)$ and $\epsilon(T) = \epsilon$. Given $\alpha > 0$:
\[
    \epsilon(t) = \epsilon \cdot \left(1 + \left(\frac{T - t}{T}\right)^\alpha\right) \enspace .      
\]

\paragraph{Infinite horizon.} Up until now, we have assumed a fixed known horizon $T$ for a recommendation session $R_T$. This is not limiting in practice: if we want to ensure P-fairness over an infinite horizon, it is enough to zero out the P-fairness constraints every $T$ users and start over for the next subsequence of $T$ users.

\section{Exact and Heuristic Solutions}

\subsection{An Integer Linear Programming Formulation} \label{sec:ilp_prog}

We first propose an Integer Linear Programming (ILP) formulation of the P-fair Bundle Recommendation Problem. To be efficient, it solves the problem over the top-$M$ items provided by a black-box single-item recommender (as mentioned in Section~\ref{sec:bundle}) instead of the entire item space $\mathcal{I}$.

\begin{align*}
    &\max_{X, Y}  \frac{1 - \gamma}{L}\sum_{i=1}^M x_{i} r_{ui} + \frac{2 \gamma}{L(L-1)} \sum_{i=1}^M \sum_{j=i+1}^M y_{ij} s_{ij}\\
    \text{s.t. } &\sum_{i=1}^M x_{i} = L\\
    & \sum_{i=1}^M x_{i} a_{iz} \le L_z, \quad \forall z=1,...,Z \\
    & \sum_{i=1}^M x_{i} (g_{ik} - \rho_k(1 - \epsilon)) + \Delta_k(1\!:\!u-1) \ge 0 \quad \forall k=1,...,K \\
    & y_{ij} \le x_i, \enspace y_{ij} \le x_j \quad \forall i=1,...,M, \quad j=i+1,...,M\\
    & y_{ij} \ge x_i + x_j - 1 \quad \forall i=1,...,M, \quad j=i+1,...,M\\
    & x_{i}, y_{ij} \in \{0,1\} \enspace.
\end{align*}

Solving the program above will provide the sequence of highest quality bundles that strictly satisfies all constraints, including the P-fairness requirements; in other words, the optimal sequence of bundles given $M$. However, this problem is NP-hard: indeed, it can be seen as a generalization of the Quadratic Knapsack Problem \cite{Gallo1980} with multiple weight constraints and a cardinality constraint, which is itself NP-hard. Therefore, it may not be possible to compute an ILP solution in reasonable time, especially if $M$ is large, which we assess in our experiments. This may limit the applicability of this approach. For these reasons, we investigate faster and more scalable approaches.


\subsection{Heuristic Solutions} \label{sec:heuristics}

Our first heuristic method, \textsc{F3R} (Alg.~\ref{alg:F3R}) (for \textsc{\underline{F}air \underline{R}andomized \underline{R}ound \underline{R}obin}), is a \textit{fairness-first} approach. \textsc{F3R} constructs the bundle iteratively by first sampling a group $G_k$ from the exposure vector distribution (line 5 to 7), and then greedily choosing the item from $G_k$ that improves the bundle score the most (line 11). Sampling according to the exposure vector ensures that the exposure targets will be met in expectation, hence \textsc{F3R} prioritizes \textit{fairness} over bundle quality. However, we can relax the fairness mechanism by choosing the best overall active item with probability $\epsilon$ (lines 8-9). $\epsilon$ therefore controls the fairness-quality trade-off. The validity constraints are then ensured, first by stopping the algorithm when the bundle size reaches $L$ (line 3), and second by removing items whose type has been covered from the active sets to meet the complementarity requirements (lines 13 to 17, subroutine referred to as \textsc{CheckComp}).

\begin{algorithm}[t]
\hspace*{\algorithmicindent} \textbf{Input:} User $u$, items \(\mathcal{I}\) \\
\hspace*{\algorithmicindent} \textbf{Output:} Bundle \(B\) 
\begin{algorithmic}[1]
    \STATE \( (I^a_1, ..., I^a_K) \gets (G_1, ..., G_K)\)  \COMMENT{Initialize active sets}
    \STATE $B = \{\}$
    \WHILE{$\vert B\vert < L$}
        \STATE $z \sim \mathcal{U}([0,1])$.
        \IF{$z > \epsilon$} 
            \STATE Sample $k \sim (\rho_1, ..., \rho_K)$
            \STATE $I^a \gets I^a_k$
        \ELSE 
            \STATE $I^a \gets \bigcup_{k=1}^K I^a_k$
        \ENDIF
        \STATE \(i \gets \text{arg}\max_{j \in I^a} \mathbf{Q}(B \cup \{j\})\)
        \STATE \(B \gets B \cup \{i\}\)           
            \FOR{\tikzmark{top} all $z$ s.t. $a_{iz} = 1$}
                \IF{$\sum_{i \in B} a_{iz} = L_z$}
                    \STATE \(I^a_k \gets I^a_k \setminus \{j \in I^a_k \vert a_{jz} = 1 \}\) for all $\ell=1,...K$ \tikzmark{right}
                \ENDIF
            \ENDFOR \tikzmark{bottom}
    \ENDWHILE
    \AddNote{top}{bottom}{right}{\textsc{CheckComp}}
    \STATE \textbf{return} \(B\)
\end{algorithmic}
\caption{\textsc{F3R}}
\label{alg:F3R}
\end{algorithm}
\begin{algorithm}[t]
\hspace*{\algorithmicindent} \textbf{Input:} User $u$, items \(\mathcal{I}\), fairness weight $\lambda$ \\
\hspace*{\algorithmicindent} \textbf{Output:} Bundle \(B\) 
\begin{algorithmic}[1]
    \STATE \(w \gets \arg \max_{i \in I} r_{ui}\)
    \STATE \(B \gets \{w\}\)
    \STATE $I^a \gets \mathcal{I} \setminus \{w\}$
    \STATE \textsc{CheckComp}
    \WHILE{$\vert B \vert < L$}
        \STATE \(i \gets \text{arg}\max_{j \in I^a} \big(Q(B \cup \{j\}) + \lambda \times \mathcal{L} (E_{k(j)}(1\!:\!u-1), \rho_{k(j)})\big)\)
        \STATE \(B \gets B \cup \{i\}\)
        \STATE \textsc{CheckComp}
    \ENDWHILE
    \STATE \textbf{return} \(B\)
\end{algorithmic}
\caption{\textsc{FairWG}}
\label{alg:FWG}
\end{algorithm}

The second method, \textsc{FairWG} (Alg.~\ref{alg:FWG}) (for \textsc{\underline{Fair}ness-\underline{W}eighted \underline{G}reedy}), is a \textit{quality-first} approach that extends the \text{BOBO-Pick} algorithm from \cite{Amer-YahiaBCFMZ14}, later taken up by \cite{BenouaretL16}. It first includes the most relevant item in the bundle, hence prioritizing  relevance (lines 1 to 3). Then, \textsc{FairWG} fills the bundle iteratively similarly to \textsc{F3R}, but augments the bundle quality function with a weighted \textit{fairness term} (line 6). In line with the problem formulation from Section~\ref{sec:cond_problem}, we define the fairness term as:
\begin{equation*}
    \mathcal{L}(E_k(1\!:\!u-1), \rho_k) = - \frac{\Delta_k(1\!:\!u-1)}{N(1\!:\!u-1)} = \rho_k - E_k(1\!:\!u-1) \enspace,
\end{equation*}
where $N(1\!:\!u-1) = \sum_{t=1}^{u-1} \vert B_t\vert$ is the total number of items recommended up to user $u-1$. This term encourages the selection of items from under-exposed groups and reduces the likelihood of choosing items from over-exposed groups.

The fairness weight $\lambda$ controls the trade-off between quality and fairness: the larger $\lambda$, the more we emphasize fairness at the detriment of bundle quality. However, it is difficult to anticipate how a specific $\lambda$ value will balance quality and fairness \textit{a priori} for a given problem, and finding the most suitable value may require several rounds of trial and error. To address this issue and make \textsc{FairWG} more widely applicable, we introduce \textsc{AdaFairWG} that dynamically adapts $\lambda := \lambda_u$ at each step $u$ based on how well the current exposures $E_k(1\!:\!u)$ align with the target exposure vector $\Rho$. 
More precisely, for a given tolerance level $\epsilon$, we aim to achieve $\mathbf{F}(B_1,...,B_T) \ge 1 - \epsilon$ through the following mechanism: if $\mathbf{F}(B_1,...,B_u) < 1 - \epsilon$, we increase the fairness weight, \textit{i.e.}, $\lambda_{u+1} = \lambda_u \times 2$, encouraging more fairness, and if $\mathbf{F}(B_1,...,B_u) > 1 - \epsilon/2$, we decrease it, \textit{i.e.}, $\lambda_{u+1} = \lambda_u / 2$, to put more emphasis on quality. With \textsc{AdaFairWG}, the recommendation provider thus only needs to specify their chosen tolerance $\epsilon$.


\section{Experiments}
\label{sec:exps}
We now empirically evaluate our methods. We examine their response time and scalability, and assess how the introduction of P-fairness affects bundle quality. 
The implementation of our methods is available at\\ \url{https://anonymous.4open.science/r/fair_bundle_reco-4231}.


\subsection{Experimental Setup} \label{sec:exp_setup}

\subsubsection{Datasets.}
We generate semi-synthetic datasets from three real-world datasets with user ratings: \textit{MovieLens-100K}
~\cite{harper2015movielens}, 
\textit{Yelp}
~\cite{yelpDataset}, 
and \textit{Amazon Electronics Retail}
~\cite{Amazon2018}. 
For each dataset, we use item features to construct relevant and realistic producer groups, compatibility functions and complementarity attributes, as defined in our bundle recommendation problem (Section~\ref{sec:prbms}). Without loss of generality, we use the SVD matrix factorization algorithm \cite{MatrixFactAlgo} to infer relevance scores from explicit user ratings. Table~\ref{tab:data_desc} summarizes our datasets along with the features used to construct the different dimensions of our problem, and we provide more details regarding dataset construction in the appendix (Section~\ref{sec:app_datasets}). 

Producer groups are formed such that a fairness-agnostic algorithm would tend to over-recommend (in a normative sense) items from some groups to the detriment of others. For instance, recommendation systems are known to be prone to popularity bias, favoring frequently rated items over less popular (or \textit{long-tail}) items. In \textit{MovieLens} and \textit{Yelp}, we categorize items into popularity tiers to promote balanced recommendations. For \textit{Amazon}, items are split on the producer's country of origin, with a majority from US-based producers. Our experiments aim to ensure a more balanced representation of items from different origins, for instance, to align with international trade requirements. 

\begin{table*}[t]
\begin{center}
\tiny
\begin{small}
\begin{tabular}{lccc}
\toprule
 & MovieLens & Yelp & Amazon \\
\midrule
Description & Movies & Restaurants & Cell Phones \& Acces.\\
\#Users & 943 & 3,838 & 64,187 \\
\#Items & 1,681 & 2,833 & 1,290 \\
\#Ratings & 99,999 & 153,362 & 78,017 \\
\midrule
Compatibility & Periode, Genre & Distance & Customer interactions \\
Producer Groups & Popularity, $K=2$ & Popularity, $K=3$ & Country, $K=4$ \\
Complementarity & \xmark & Type (\textit{e.g.}, Italian) & Type (\textit{e.g.}, Headphones) \\
\bottomrule
\end{tabular}
\end{small}
\end{center}
\caption{Description of the datasets.}
\label{tab:data_desc}
\end{table*}

\subsubsection{Methods and Implementation.}

We assess our three recommendation methods: \textsc{ILP}, \textsc{F3R} and \textsc{FairWG}. All methods are implemented in Python and experiments are run on Intel i7 CPUs. For \textsc{ILP}, we implement the program from Section~\ref{sec:ilp_prog} using the Gurobi solver.
We also use the \textsc{ILP} method without P-fairness (or, equivalently, with tolerance $\epsilon=1.0$) as a baseline. Indeed, given the $M$ most relevant items, we can build the highest-quality valid bundle for each user. We then compare this "optimal" bundle with the fair bundle returned by \textsc{ILP}, \textsc{F3R} and \textsc{FairWG}, to evaluate the impact of P-fairness on quality. We also compare them with a random baseline to ensure that our methods indeed provide meaningful recommendations.

We further explore the capabilities of our methods by varying the different problem parameters.
We consider two bundle sizes, \textit{Small} ($L=6$) and \textit{Large} ($L=12$), and --- unless stated otherwise --- set $\gamma=1/3$.
We also vary algorithm-specific parameters $\alpha$ (\textsc{ILP}), $\epsilon$ (\textsc{ILP}, \textsc{F3R}) and $\lambda$ (\textsc{FairWG}). 
Our results are averages of five recommendation sessions with different random seeds, and we provide confidence intervals of the form $[m-\sigma, m+\sigma]$, where $m$ is the mean and $\sigma$ the standard deviation. Since \textsc{F3R} introduces additional randomness, we also run five random seeds for each recommendation session for this algorithm.

\subsubsection{Measures.} 
\underline{\textit{Quality:}} the quality of a bundle (a combination of relevance and compatibility) as given by the bundle scoring function $\mathbf{Q}(B)$. We use the term \textit{Relative Quality} to refer to the ratio $\mathbf{Q}(B) / \mathbf{Q}(B^\star)$, where $B^\star$ denotes the non-fair "optimal" bundle obtained with the fairness-free \textsc{ILP} program (for the same user and same $M$).
\underline{\textit{P-Fairness:}} Producer-fairness ranges from 0 to 1 and is equal to 1 in case of perfect fairness, \textit{i.e.}, if $E_k = \rho_k$:
    $
       ~ \mathbf{F}(B_1,...,B_T) = 1 - \max_{k=1,..,K} \max\left(0, (\rho_k - E_k) / \rho_k\right)).
    $
    It is built such that, if we allow a tolerance $\epsilon$, we expect $\mathbf{F}(B_1,...,B_T) \ge 1 - \epsilon$. 
\underline{\textit{Runtime:}} measured as response time per user, a key component of user satisfaction in interactive recommendation settings.

\subsection{Empirical Findings}

\subsubsection{Proposed methods enhance recommendation fairness over fairness-agnostic baseline.}

\begin{figure}[t]
\centering
\begin{minipage}{.48\textwidth}
\includegraphics[width=1.0\linewidth]{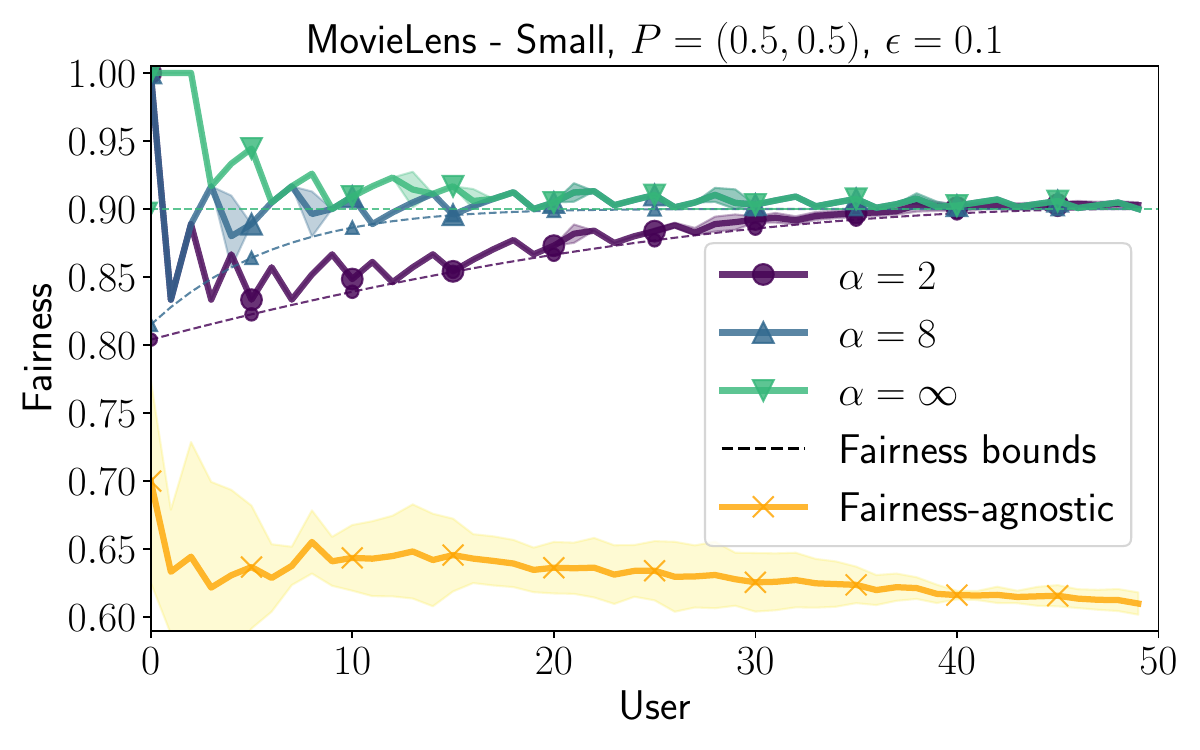}

\caption{Fairness evolution in a session for \textsc{ILP} on \textit{MovieLens} ($\epsilon=10\%$).}
\label{fig:ml_ilp_results_alpha}
\end{minipage}
\hspace{0.2cm}
\begin{minipage}{.48\textwidth}
\begin{center}
\tiny
\begin{tabular}{@{}l@{\hspace{12pt}}c@{\hspace{12pt}}c@{\hspace{12pt}}c@{}}
\toprule
\multicolumn{4}{c}{\textbf{ILP}} 
\\ \midrule
        $M$      & MovieLens                   & Yelp             & Amazon           \\  \midrule
$50$ & $0.68 \pm 0.29$ & $1.03 \pm 0.37$ &       $0.14 \pm 0.04$     \\
$100$    & $3.39 \pm2.74$  & $12.85 \pm 3.80$    &     $0.46 \pm 0.11$            \\
$200$    & $88.61 \pm 23.24$ & $184.03 \pm 126.99$ &        $2.67 \pm 0.44$             \\ \midrule
\multicolumn{4}{c}{\textbf{FairWG}, \textbf{F3R}} 
\\ \midrule
       $M$      & MovieLens                   & Yelp             & Amazon           \\  \midrule
$50$ & \multirow{4}{*}{$<10^{-2}$} & \multirow{4}{*}{$<10^{-2}$} & \multirow{4}{*}{$<10^{-2}$}\\ 
$100$ & & & \\
$200$ & & & \\
$500$ & & & \\
 \bottomrule
\end{tabular}
\end{center}
\caption{\textsc{ILP}, \textsc{FairWG}, and \textsc{F3R} runtimes per user (in seconds) for different values of $M$.}
\label{tab:ilp_runtime_main}
\end{minipage}
\end{figure}

As illustrated in Fig.~\ref{fig:ml_ilp_results_alpha} (where the fairness target is met when $\mathbf{F} \ge 0.95$), ILP consistently achieves the fairness objective by enforcing the P-fairness constraint at each step. Indeed, $\mathbf{F}$ (plain lines) never breaches the imposed bound (dashed lines). In contrast, the fairness-agnostic baseline delivers a very unfair solution.
Table~\ref{tab:bundle_example} illustrates a \textit{MovieLens} bundle from our ILP solution, with and without P-fairness awareness.
To increase exposure for the least popular movie group $G_1$, the fair solution includes \textit{A Time to Kill} and \textit{Some Folks Call It a Slim Blade}, which are nearly as relevant as the more popular alternatives (\textit{Hamlet} and \textit{Good Hunting}) selected without the P-fairness constraint. 
This selection enhances $G_1$'s exposure and meets P-fairness requirements.
Although not ensuring strict P-fairness like \textsc{ILP}, our heuristic approaches \textsc{FairWG} and \textsc{F3R} successfully provide fairer recommendations than the baseline (\textit{i.e.}, the ILP method with tolerance $\epsilon=1.0$), as demonstrated in Fig.~\ref{fig:ml_all_methods}. Results are consistent across all datasets (see also Fig.~\ref{fig:all_trade_offs_yelp_amazon} in the Appendix).


\begin{table}[t]
\begin{center}
\begin{small}
\begin{tabular}{lccc@{\hspace{5pt}}ccc}
\toprule
& \multicolumn{3}{c}{Optimal Bundle ($\textbf{R} = 0.90$, $\textbf{S}=0.92$)} & \multicolumn{3}{c}{Fair Bundle ($\textbf{R} = 0.88$, $\textbf{S}=0.84$)}  \\
\midrule
& Movie Title & \# votes & $r_{ui}$ & Title & \# votes & $r_{ui}$ \\
\midrule
$i_1$ & Madness of King George & 19K & 0.85 & Madness of King George & 19K & 0.85 \\
$i_2$ & A Bronx Tale & 167K & 0.87 & A Bronx Tale & 167K & 0.87 \\
$i_3$ & \textbf{Hamlet} & \textbf{41K} &\textbf{ 0.88} & \textbf{A Time to Kill} & \textbf{1K} & \textbf{0.83} \\
$i_4$ & Searching for Bobby F... & 43K & 0.90 & Searching for Bobby F... & 43K & 0.90 \\
$i_5$ & \textbf{Good Will Hunting} & \textbf{1,100K} & \textbf{0.93} & \textbf{Some Folks Call It...} & \textbf{2K} & \textbf{0.87}\\
$i_6$ & Shawshank Redemption & 2,970K & 0.96 & Shawshank Redemption & 2,970K & 0.96 \\
\bottomrule
\end{tabular}
\end{small}
\end{center}
\caption{Example of a bundle (\textit{MovieLens}, user $u=486$) produced by our recommendation algorithm: the Optimal Bundle (\textit{left}) is returned by \textsc{ILP} \textbf{without} the fairness constraint and the Fair Bundle (\textit{right}) is returned by \textsc{ILP}  \textbf{with} the fairness constraint (with $\epsilon=0$). Items in \textbf{bold} reflects differences between the Optimal and  Fair solutions.}
\label{tab:bundle_example}
\end{table}


\subsubsection{Introducing P-fairness adversely impacts quality, revealing a trade-off.}

\begin{figure}[tt]
\centering
\begin{minipage}{.49\textwidth}
\includegraphics[width=\linewidth]{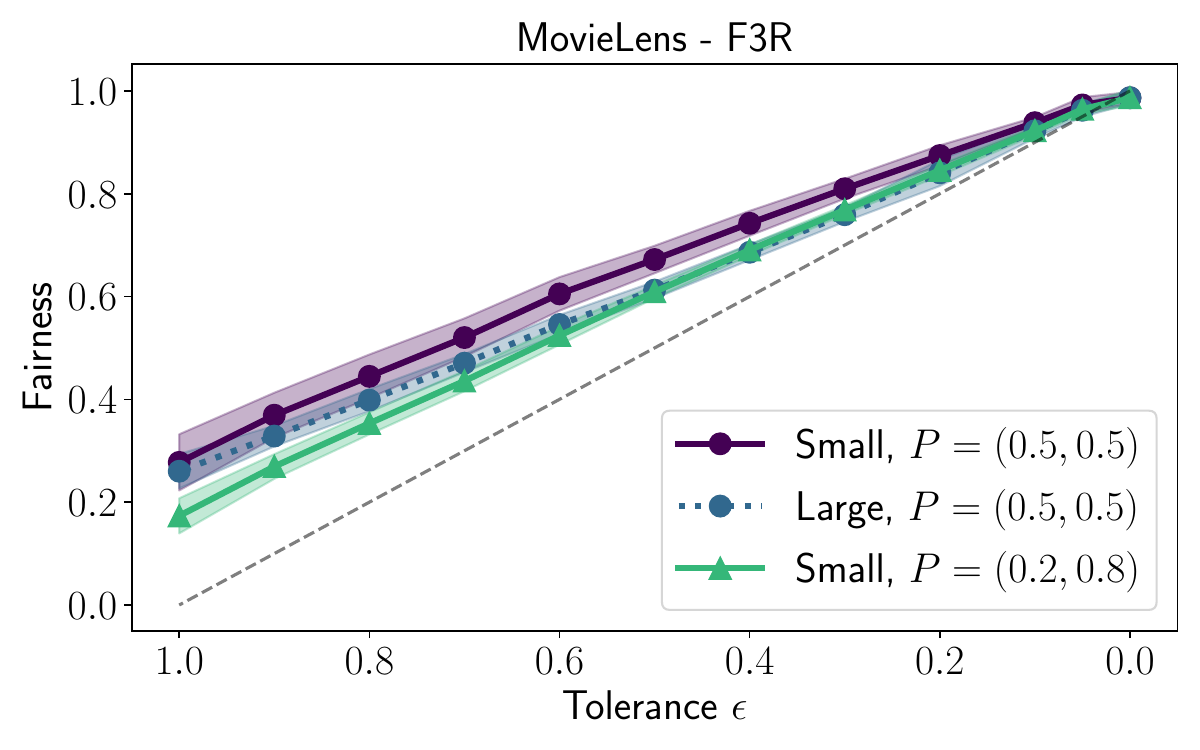}
\end{minipage}
\begin{minipage}{.49\textwidth}
\includegraphics[width=\linewidth]{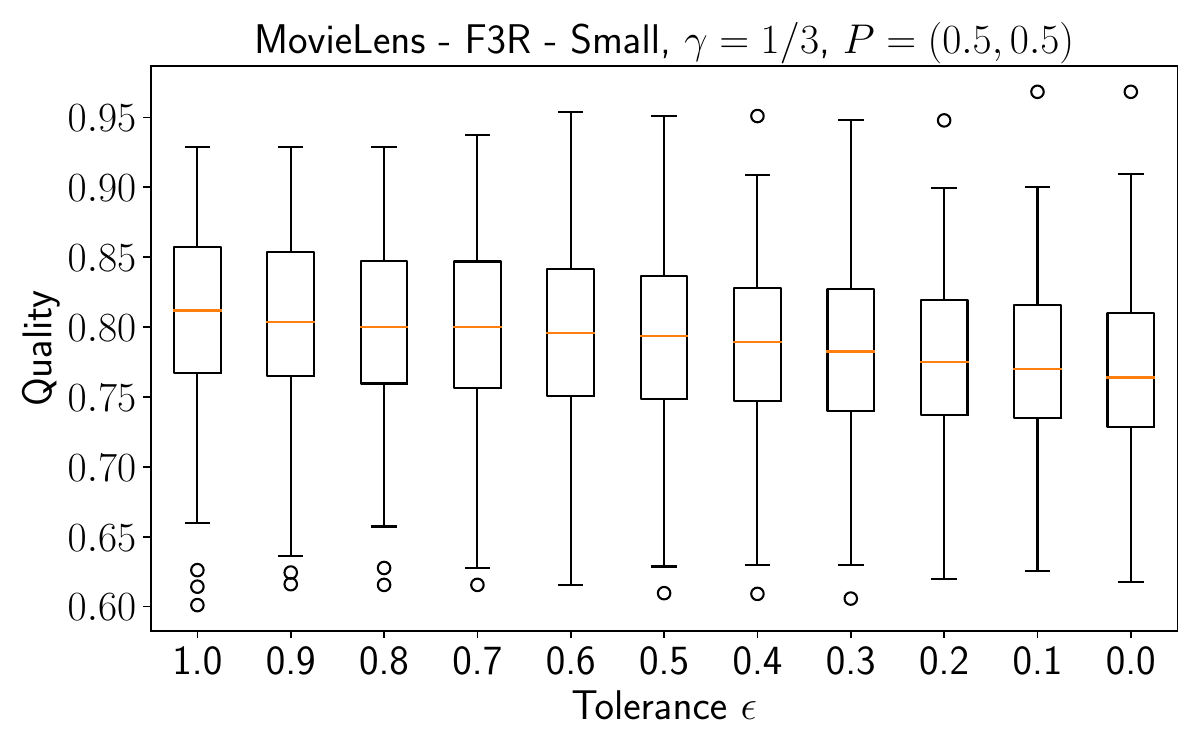}
\end{minipage}
\begin{minipage}{.49\textwidth}
\includegraphics[width=\linewidth]{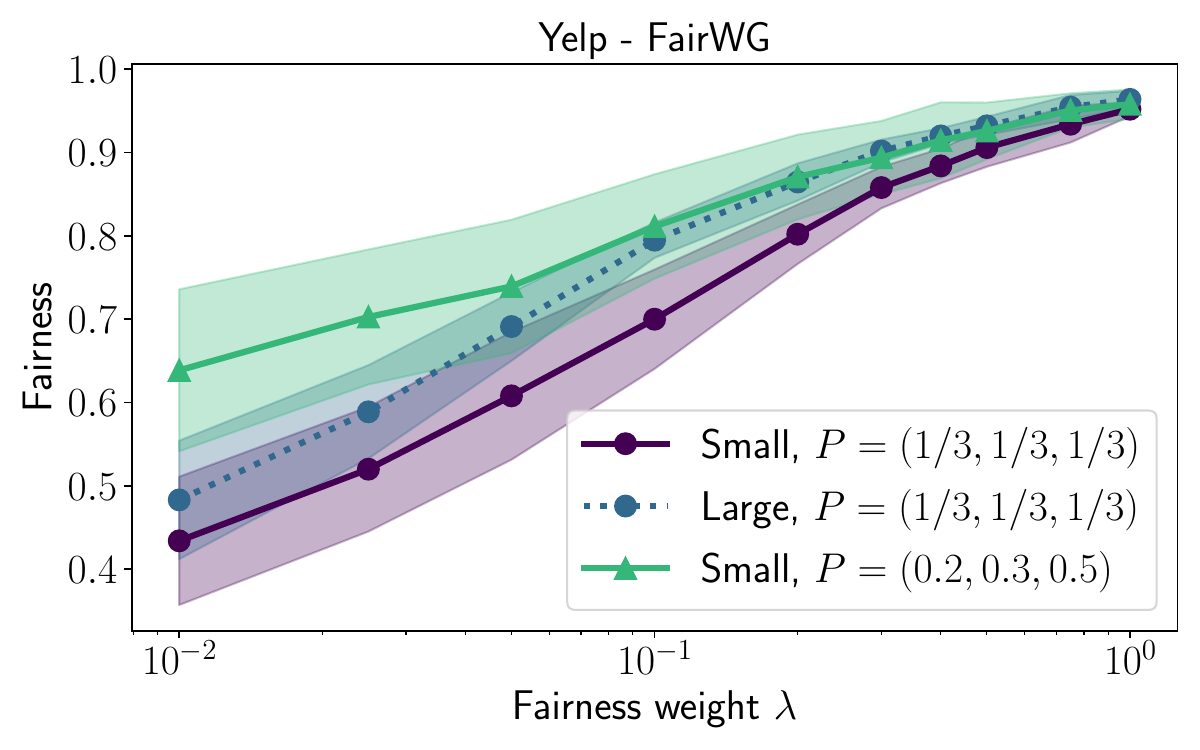}
\end{minipage}
\begin{minipage}{.49\textwidth}
\includegraphics[width=\linewidth]{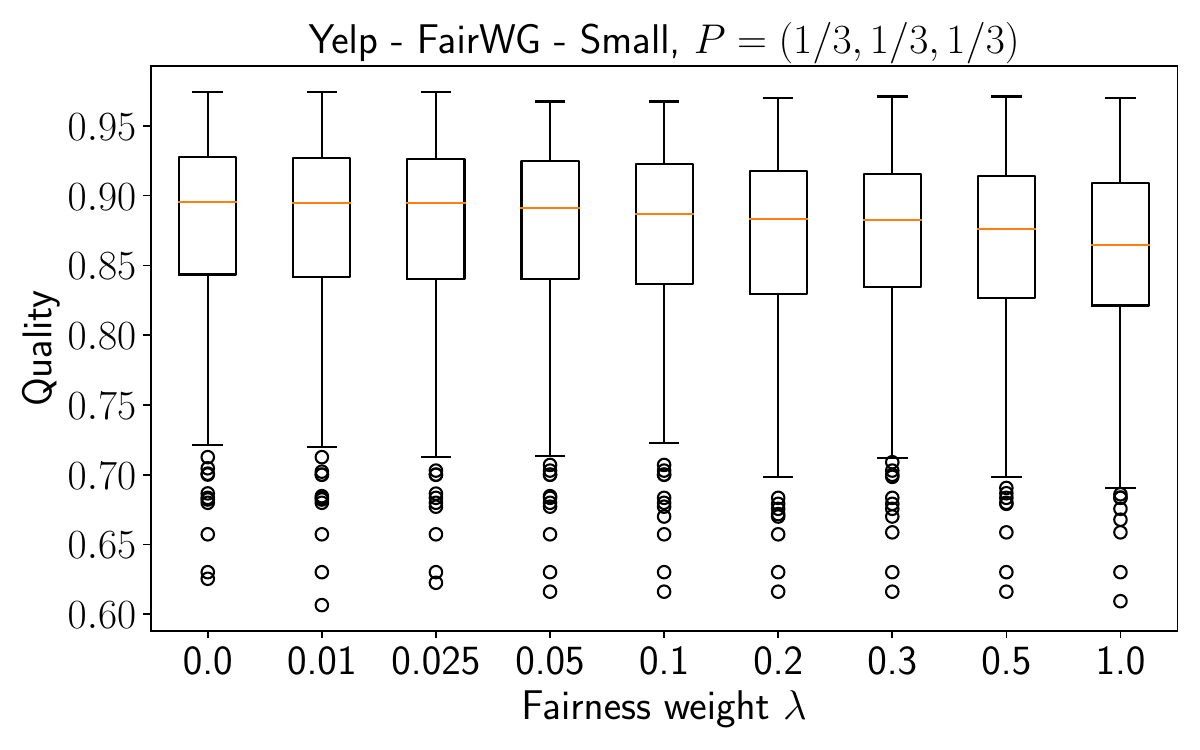}
\end{minipage}
\caption{Fairness (\textit{left}) and bundle quality (\textit{right}) for heuristic methods. The fairness-quality trade-off is achieved with $\epsilon$ for \textsc{F3R} and with $\lambda$ for \textsc{FairWG}.}
\label{fig:ml_heuristics}
\end{figure}

Fig.~\ref{fig:ml_heuristics} and Fig.~\ref{fig:ml_all_methods} show that the introduction of P-fairness comes at a cost and that one generally cannot obtain a solution that is simultaneously more fair and of higher quality. In fact, for all three methods, increasing the fairness metric leads to a decrease in quality. For \textsc{ILP} and \textsc{F3R}, this quality-fairness trade-off is controlled by the tolerance parameter $\epsilon \in (0, 1)$, whereas for \textsc{FairWG}, the trade-off is controlled by the weight $\lambda \ge 0$: by increasing $\lambda$ or decreasing $\epsilon$, we put more emphasis on fairness to the detriment of quality. Similar conclusions are drawn on the other datasets (Fig.~\ref{fig:all_trade_offs_yelp_amazon} and Fig.~\ref{fig:f3r_yelp}-\ref{fig:fbp_amazon} in the Appendix).


\subsubsection{\textsc{ILP} generally achieves the best quality-fairness trade-off for small instances.}

\begin{figure}[t]
\centering
\begin{minipage}{.48\textwidth}
\includegraphics[width=1.05\linewidth]{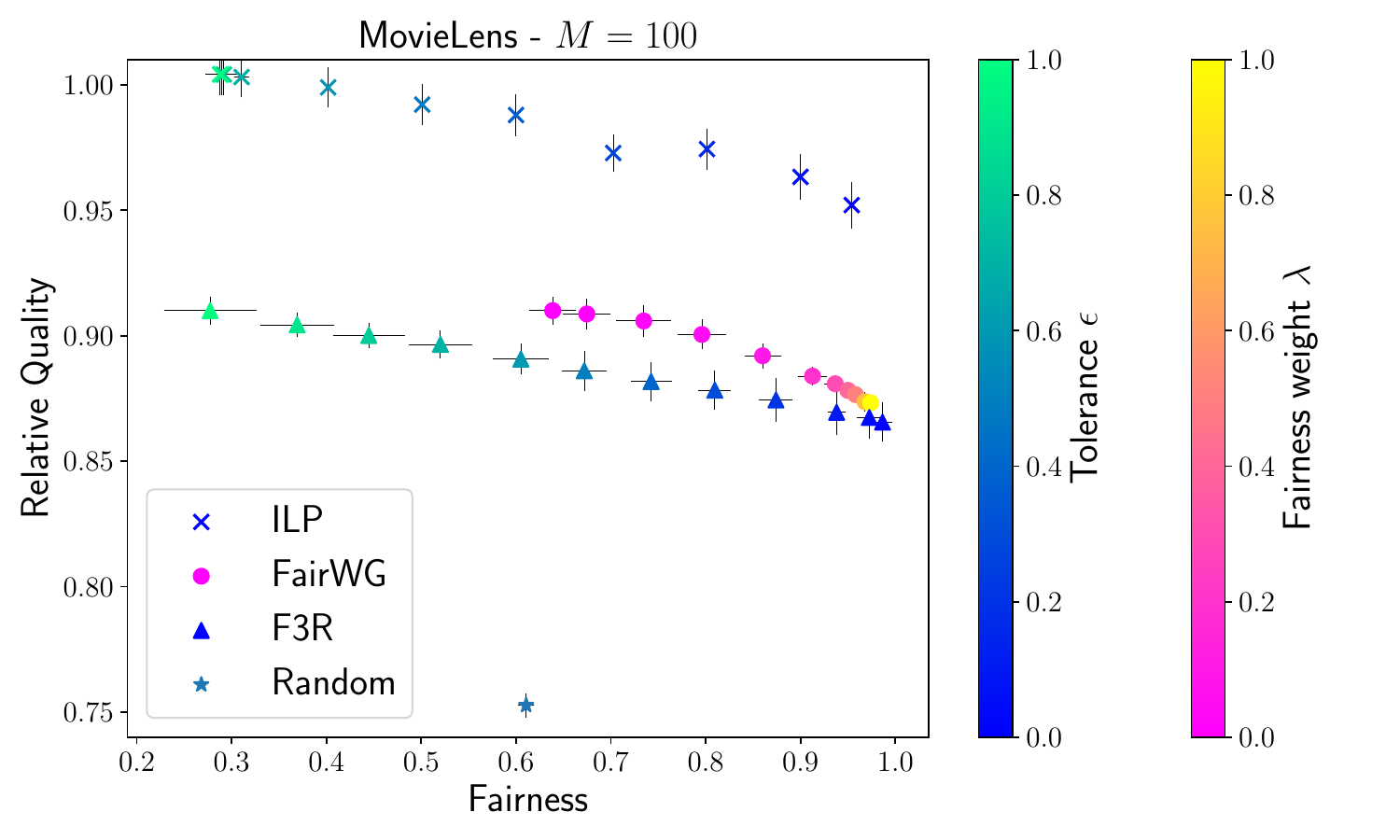}
\caption{Fairness-utility trade-offs on \textit{MovieLens} with $M=100$. Quality is relative to the optimal bundle produced by \textsc{ILP} w/o P-fairness (\textit{i.e.}, with $\epsilon = 1.0$).}
\label{fig:ml_all_methods}
\end{minipage} \hspace{0.2cm}
\begin{minipage}{.48\textwidth}
\includegraphics[width=1.05\linewidth]{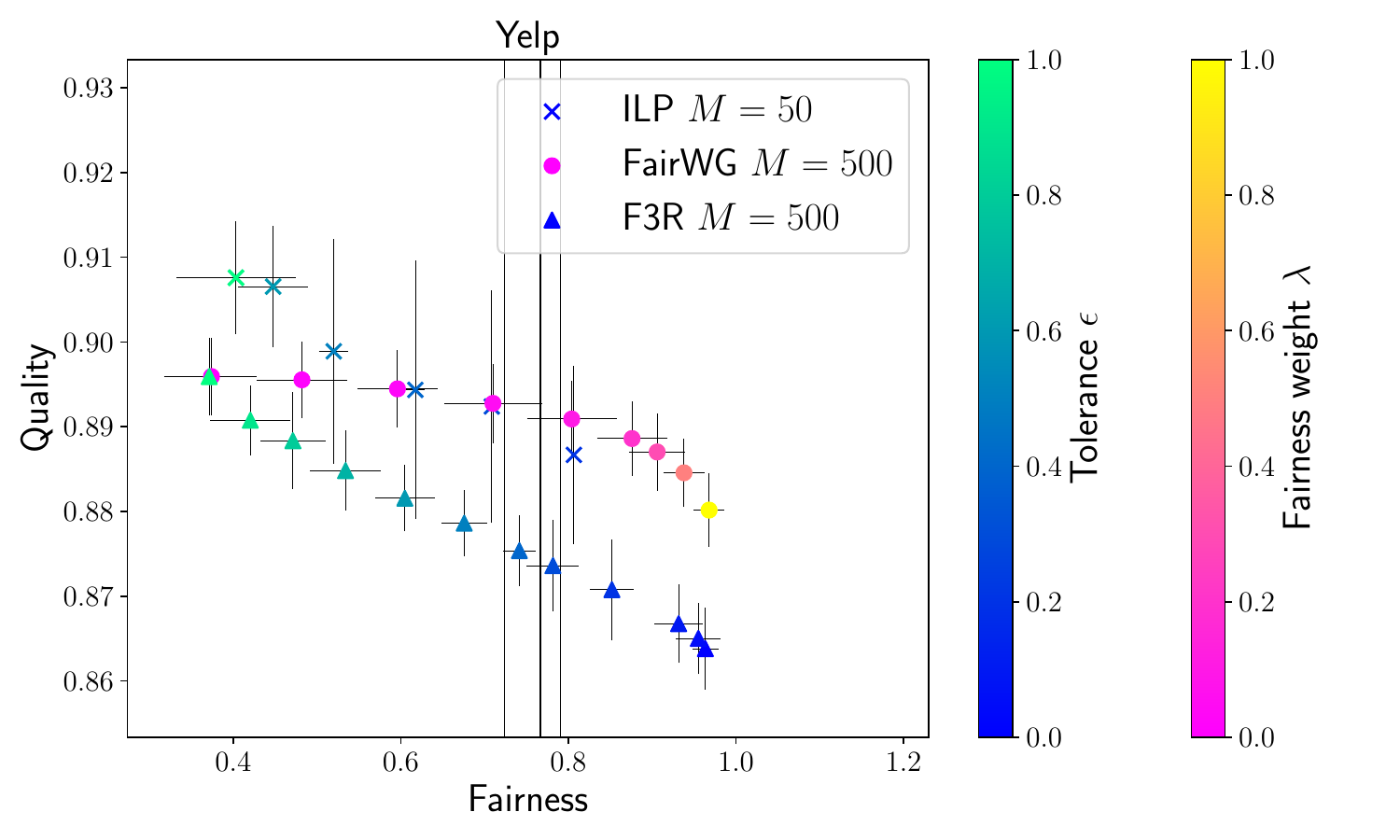}
\caption{Fairness-utility trade-offs on \textit{Yelp} for comparable runtimes ($\approx1s$/user for \textsc{ILP} with $M=50$, $\lessapprox 10^{-2}s$/user for \textsc{FairWG}, \textsc{F3R} with $M=500$).}
\label{fig:ilp_heuristic_comp_sameruntime}
\end{minipage}
\end{figure}

\begin{figure}[ht]
\centering
\begin{minipage}{.49\textwidth}
\includegraphics[width=\linewidth]{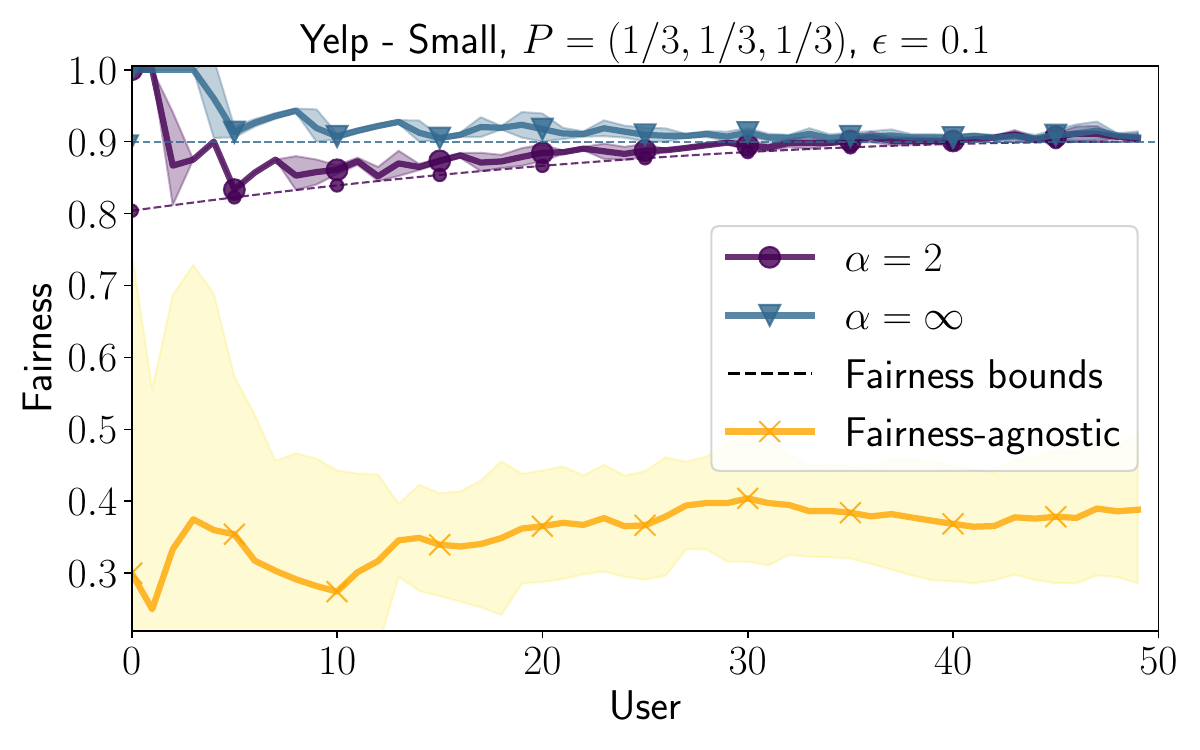}
\end{minipage}
\begin{minipage}{.49\textwidth}
\includegraphics[width=\linewidth]{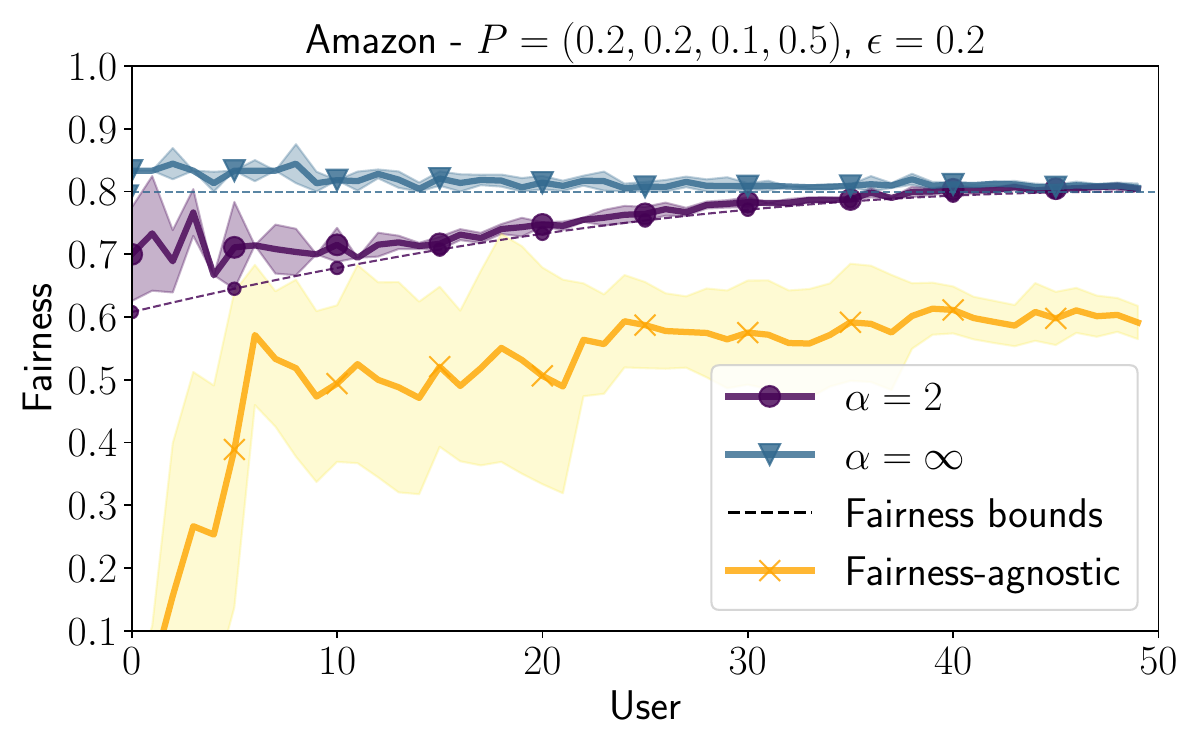}
\end{minipage}
\caption{Results for ILP on \textit{Yelp} (\textit{left}) and \textit{Amazon} (\textit{right}): fairness metric across a recommendation session for different values of $\alpha$.}
\label{fig:fairness_alpha_yelp_amazon}
\end{figure}

Fig.~\ref{fig:ml_all_methods} and Fig.~\ref{fig:all_trade_offs_yelp_amazon} show that, for a fixed $M$, \textsc{ILP} yields superior quality trade-offs compared to the two proposed heuristics. However, this is contingent on the feasibility of the problem. On more complex problems involving complementarity constraints (\textit{Amazon} and \textit{Yelp}), \textsc{ILP}  fails to provide solutions for low tolerance values and is therefore unable to reach high levels of P-fairness. This is particularly noticeable on \textit{Amazon} where the program is not solved for $\epsilon < 0.2$. In some of these difficult cases, using adaptive exposure vectors (illustrated in Fig.~\ref{fig:ml_ilp_results_alpha} and Fig.~\ref{fig:fairness_alpha_yelp_amazon}) can help reach solutions by relaxing the fairness constraint in the earlier stages of the recommendation session. In Table~\ref{tab:ilp_alpha}, we indeed observe that the ILP program fails in several problem instances with $\alpha=\infty$ (\textit{i.e.}, fixed exposure vector) but can recover high-quality solutions with $\alpha=2$ (\textit{i.e.}, adaptive exposure vector). Moreover, \textsc{ILP} is by far the slowest approach (as shown in Fig.~\ref{tab:ilp_runtime_main}), and despite the increase in quality associated with larger $M$, it has to remain small if we consider deploying it in real-time settings (see Table~\ref{tab:ilp_runtime}). 

\subsubsection{Heuristics achieve competitive trade-offs at minimal cost compared to ILP.}

While slightly inferior to \textsc{ILP} for a given $M$, \textsc{F3R} and \textsc{FairWG} still provide high quality bundles while encouraging fairness (Fig.~\ref{fig:ml_all_methods}). 
As opposed to \textsc{ILP}, these heuristics also enjoy a near-constant runtime with respect to $M$ (Fig.~\ref{tab:ilp_runtime_main}). Increasing $M$ can lead to significant improvements on quality (see Table~\ref{tab:ilp_runtime} in the Appendix). Therefore, if we contrast all methods for comparable runtimes instead of a fixed $M$, as done in Fig.~\ref{fig:ilp_heuristic_comp_sameruntime} (as well as Fig.~\ref{fig:ilp_heuristic_comp_sameruntime_other_dataset} in the Appendix), we observe that they are actually very close in performance. In fact, \textsc{FairWG} can even achieve better quality-fairness trade-offs than \textsc{ILP} at high fairness levels. 

\subsubsection{While \textsc{F3R} can target specific fairness levels, it is generally outperformed by \textsc{FairWG}. Meanwhile, \textsc{AdaFairWG} successfully addresses the limitations of \textsc{FairWG}.}

\begin{figure}[tt]
\centering
\begin{minipage}{.49\textwidth}
\includegraphics[width=\linewidth]{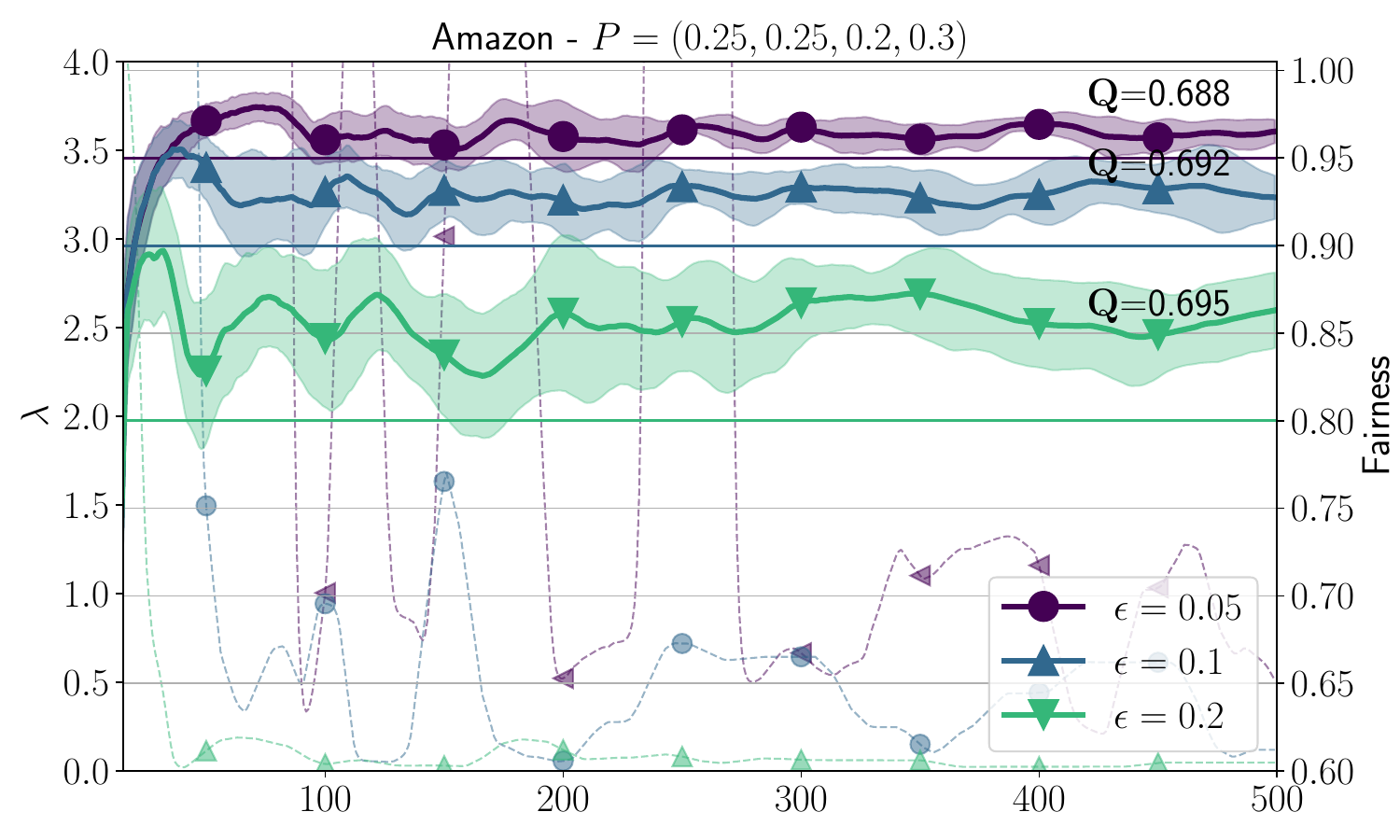}
\end{minipage}
\begin{minipage}{.49\textwidth}
\includegraphics[width=\linewidth]{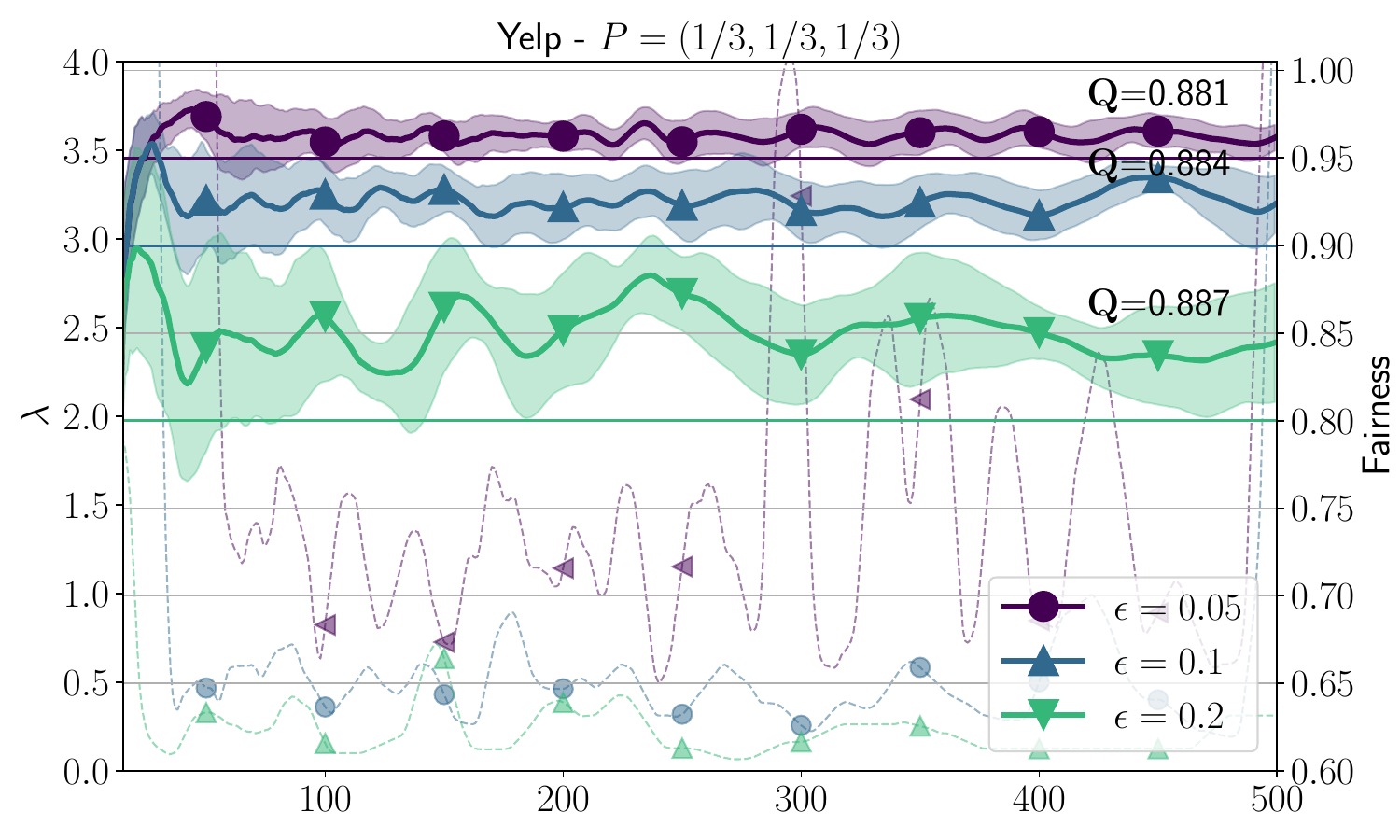}
\end{minipage}
\caption{Results for \textsc{AdaFairWG} with tolerance levels $\epsilon \in \{5\%, 10\%, 20\%\}$ and horizon $T=500$, for \textit{Amazon} (\textit{left}) and \textit{Yelp} (\textit{right}). Plain lines: fairness $\mathbf{F}$. Dashed lines: adaptive fairness weight $\lambda$. Beside each curve, the average quality $\mathbf{Q}$ is displayed. }
\label{fig:adafairwg}
\end{figure}

\begin{figure}[ht]
\centering
\begin{minipage}{.49\textwidth}
\includegraphics[width=\linewidth]{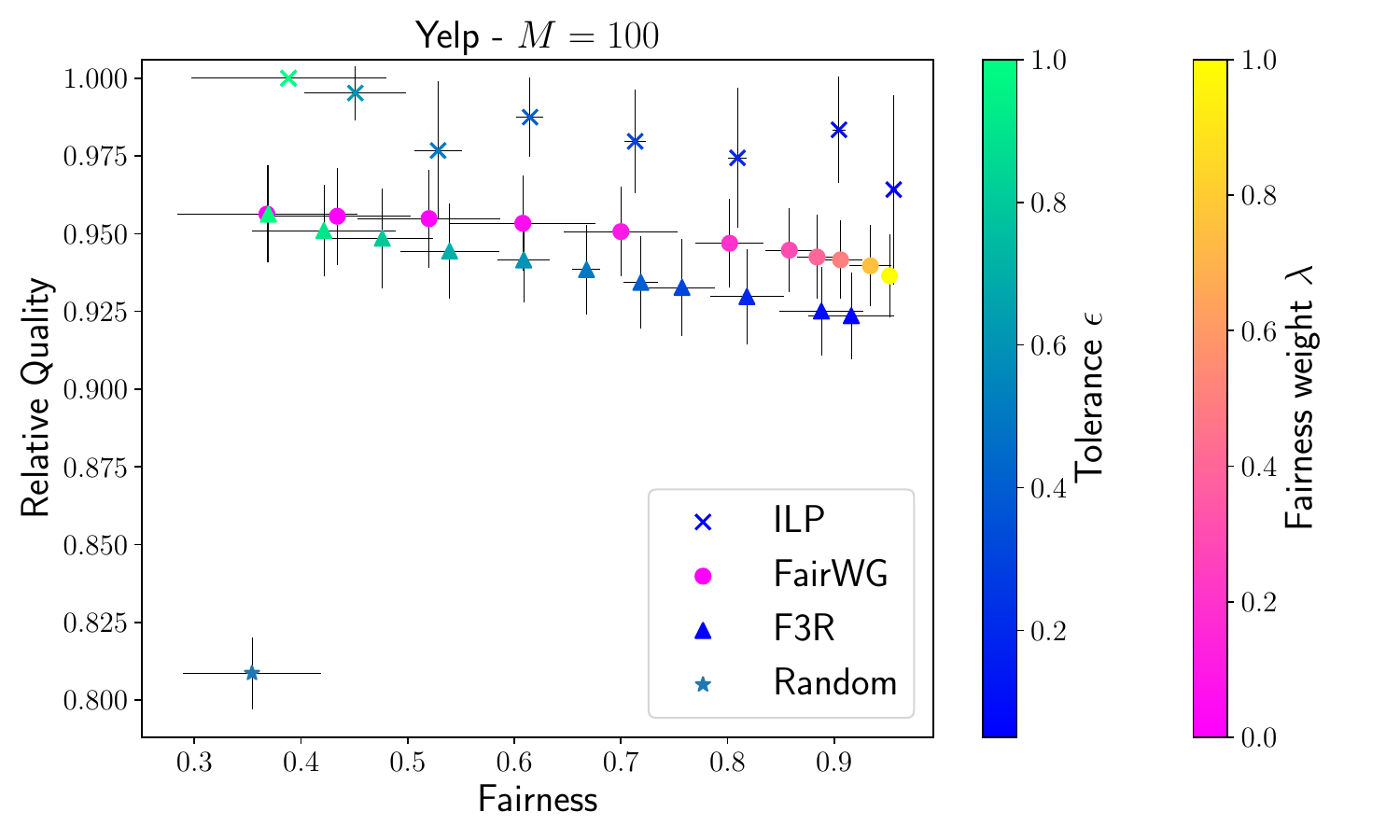}
\end{minipage}
\begin{minipage}{.49\textwidth}
\includegraphics[width=\linewidth]{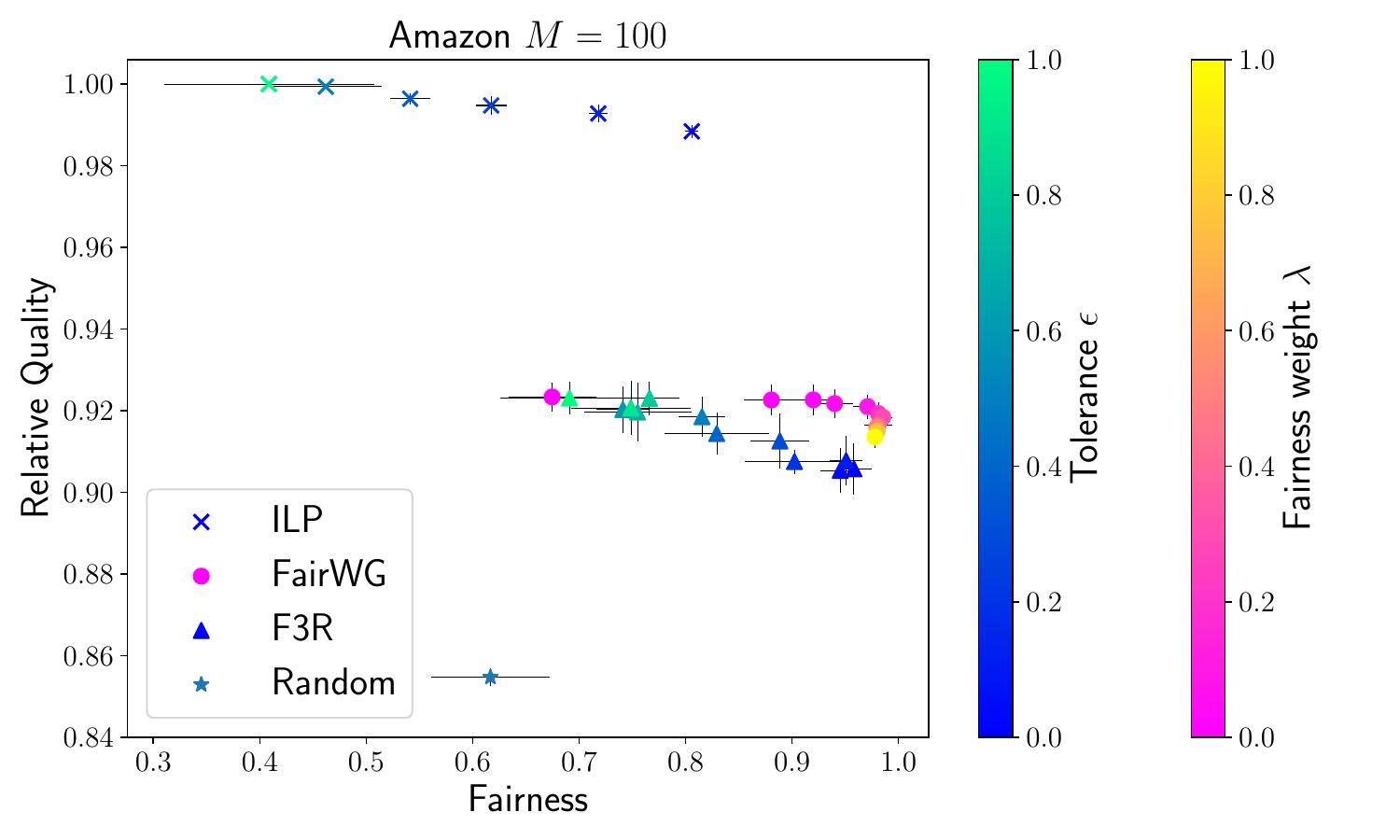}
\end{minipage}
\caption{Fairness-quality trade-offs for all methods on Yelp (\textit{left}) and Amazon (\textit{right}). For \textsc{ILP}, low $\epsilon$'s for which solutions cannot be reached are not represented. 
}
\label{fig:all_trade_offs_yelp_amazon}
\end{figure}

Given a tolerance level $\epsilon$, \textsc{F3R} generally achieves the fairness goal $\mathbf{F}(B_1,...,B_T) \ge 1 - \epsilon$ as expected (see Fig.~\ref{fig:ml_heuristics} as well as Fig.~\ref{fig:f3r_yelp} and \ref{fig:f3r_amazon} in the Appendix). However, it might not meet this objective for $\epsilon$ close to zero, especially in the more difficult \textit{Yelp} and \textit{Amazon} problems (see Fig.~\ref{fig:all_trade_offs_yelp_amazon}, \ref{fig:f3r_yelp} and \ref{fig:f3r_amazon}). Indeed, when fairness is hard to achieve, there may not be enough available items in each group $G_k$ to keep sampling according to the exposure vector distribution until the bundle is filled, resulting in a mismatch between the achieved and the expected fairness level. This can be alleviated by increasing $M$, expanding the range of available items, as can be seen in the \textit{Amazon} experiments. 
On the other hand, \textsc{FairWG} generally performs better, delivering higher quality recommendations across all fairness levels (see Fig.~\ref{fig:ml_all_methods}, Fig.~\ref{fig:all_trade_offs_yelp_amazon} and Fig.~\ref{fig:ilp_heuristic_comp_sameruntime_other_dataset} in the Appendix), but requires choosing $\lambda$ through trial and error.
As shown in Fig.~\ref{fig:adafairwg}, \textsc{AdaFairWG}, which dynamically adjusts $\lambda$ (in \textit{dashed lines}) during the session, overcomes this difficulty and can successfully achieve a given target tolerance level $\epsilon$. Using a large horizon $T=500$, we observe that the fairness level ends up stabilizing just over the target level. This suggests that this method is highly effective in controlling the fairness level over large horizons.

\section{Related Work}
\textbf{Bundle Recommendation.}
A number of works have investigated the recommendation of sets of items, as opposed to traditional single-item recommendation. 
Two main lines of research emerge from the literature: recommendation of existing bundles (by exploiting user-item and user-bundle historical interactions \cite{ChangG0JL20gcn_rec,LiLWSXYYZL21attention,MultiCBR,SunFYFQOL24revisiting}) and bundle generation (by automatically building relevant sets of items). While some approaches create bundles based purely on user past interactions, including existing bundles or co-purchases \cite{BaiZSQALG19,Chang23,Hu0GZ20modeling}, our work is closer to \cite{Amer-YahiaBCFMZ14,BenouaretL16,XieLW12} which view bundles as sets of relevant, compatible items satisfying some validity constraints, and do not assume predefined bundles.
\textit{As far as we know, no work has addressed fairness in (sequential) bundle recommendation.}

\noindent\textbf{Fairness of Exposure in Recommendation.}
Achieving fairness of exposure has been approached at different stages of a recommendation process: pre-processing \cite{chen2023improving,fang2022fairroad,rastegarpanah2019fighting} such as interventions on input data to achieve a fairness goal~\cite{DBLP:conf/sigmod/SalimiRHS19}, in-processing \cite{burke2018balanced,yao2017beyond} such as exploring different weights of a scoring function to achieve a fair ranking~\cite{DBLP:conf/sigmod/AsudehJS019}, or post-processing where data is re-ranked~\cite{li2021user,CPFair}. In sequential settings, the notion of amortized fairness has been proposed as a post-processing where output recommendations are re-sorted across a sequence of queries~\cite{BiegaGW18}. {\em While our work belongs to in-processing, none of the existing proposals applies to high quality sequential bundle recommendation.}

\section{Conclusion}

In this work, we investigated P(roducer)-fairness in the context of sequential bundle recommendation. To the best of our knowledge, our work is the first to combine quality and fairness in bundle recommendation.
The need to account for multiple dimensions, pairwise item interactions, and P-fairness across the entire recommendation sequence makes this problem challenging from both formalization and computational perspectives.
We formulate the P-fair bundle recommendation problem in an expressive and feasible manner and propose both exact and heuristic solutions. 
Empirical investigations show that our methods successfully enhance recommendation fairness in bundles and highlight the strengths and limitations of each approach. In particular, we show that our heuristic algorithms can be as effective as the expensive ILP solution, especially when dynamically balancing fairness and quality during the recommendation session.

Our work opens a number of new avenues in research. An immediate one is to study the applicability of other notions of exposure in the context of bundle recommendation~\cite{SinghJ18}. 
Another avenue is to formalize and solve fairness of ranking bundles~\cite{DBLP:conf/sigmod/RoyACDY10}.
Finally, it would also be interesting to measure the impact of introducing P-fairness on individual users, \textit{i.e.}, consumer-fairness, as in \cite{CPFair}.

%
%
%
\bibliographystyle{splncs04}
\bibliography{main}

\begin{thebibliography}{10}
\providecommand{\url}[1]{\texttt{#1}}
\providecommand{\urlprefix}{URL }
\providecommand{\doi}[1]{https://doi.org/#1}

\bibitem{Amer-YahiaBCFMZ14}
Amer{-}Yahia, S., Bonchi, F., Castillo, C., Feuerstein, E., M{\'{e}}ndez{-}D{\'{\i}}az, I., Zabala, P.: Composite retrieval of diverse and complementary bundles. {IEEE} Trans. Knowl. Data Eng.  \textbf{26}(11),  2662--2675 (2014)

\bibitem{DBLP:conf/sigmod/AsudehJS019}
Asudeh, A., Jagadish, H.V., Stoyanovich, J., Das, G.: Designing fair ranking schemes. In: {SIGMOD}. pp. 1259--1276 (2019)

\bibitem{BaiZSQALG19}
Bai, J., Zhou, C., Song, J., Qu, X., An, W., Li, Z., Gao, J.: Personalized bundle list recommendation. In: The World Wide Web Conference. pp. 60--71 (2019)

\bibitem{BenouaretL16}
Benouaret, I., Lenne, D.: A package recommendation framework for trip planning activities. In: RecSys. pp. 203--206 (2016)

\bibitem{BiegaGW18}
Biega, A.J., Gummadi, K.P., Weikum, G.: Equity of attention: Amortizing individual fairness in rankings. In: {SIGIR}. pp. 405--414 (2018)

\bibitem{burke2018balanced}
Burke, R., Sonboli, N., Ordonez-Gauger, A.: Balanced neighborhoods for multi-sided fairness in recommendation. In: Conference on fairness, accountability and transparency. pp. 202--214. PMLR (2018)

\bibitem{ChangG0JL20gcn_rec}
Chang, J., Gao, C., He, X., Jin, D., Li, Y.: Bundle recommendation with graph convolutional networks. In: {SIGIR}. pp. 1673--1676 (2020)

\bibitem{Chang23}
Chang, J., Gao, C., He, X., Jin, D., Li, Y.: Bundle recommendation and generation with graph neural networks. {IEEE} Trans. Knowl. Data Eng.  \textbf{35}(3),  2326--2340 (2023)

\bibitem{chen2023improving}
Chen, L., Wu, L., Zhang, K., Hong, R., Lian, D., Zhang, Z., Zhou, J., Wang, M.: Improving recommendation fairness via data augmentation. arXiv preprint arXiv:2302.06333  (2023)

\bibitem{fang2022fairroad}
Fang, M., Liu, J., Momma, M., Sun, Y.: Fairroad: Achieving fairness for recommender systems with optimized antidote data. In: Proceedings of the 27th ACM on Symposium on Access Control Models and Technologies. pp. 173--184 (2022)

\bibitem{Gallo1980}
Gallo, G., Hammer, P.L., Simeone, B.: Quadratic knapsack problems, pp. 132--149. Springer Berlin Heidelberg (1980)

\bibitem{harper2015movielens}
Harper, F.M., Konstan, J.A.: The movielens datasets: History and context. ACM Transactions on Interactive Intelligent Systems  \textbf{5}(4),  1--19 (2015)

\bibitem{Hu0GZ20modeling}
Hu, H., He, X., Gao, J., Zhang, Z.: Modeling personalized item frequency information for next-basket recommendation. In: {SIGIR}. pp. 1071--1080 (2020)

\bibitem{LiLWSXYYZL21attention}
Li, C., Lu, Y., Wang, W., Shi, C., Xie, R., Yang, H., Yang, C., Zhang, X., Lin, L.: Package recommendation with intra- and inter-package attention networks. In: {SIGIR}. pp. 595--604 (2021)

\bibitem{li2021user}
Li, Y., Chen, H., Fu, Z., Ge, Y., Zhang, Y.: User-oriented fairness in recommendation. In: Proceedings of the web conference 2021. pp. 624--632 (2021)

\bibitem{MultiCBR}
Ma, Y., He, Y., Wang, X., Wei, Y., Du, X., Fu, Y., Chua, T.: Multicbr: Multi-view contrastive learning for bundle recommendation. {ACM} Trans. Inf. Syst.  \textbf{42}(4),  100:1--100:23 (2024)

\bibitem{MehrotraMBL018}
Mehrotra, R., McInerney, J., Bouchard, H., Lalmas, M., Diaz, F.: Towards a fair marketplace: Counterfactual evaluation of the trade-off between relevance, fairness {\&} satisfaction in recommendation systems. In: {CIKM}. pp. 2243--2251 (2018)

\bibitem{CPFair}
Naghiaei, M., Rahmani, H.A., Deldjoo, Y.: {CPFair: Personalized Consumer and Producer Fairness Re-ranking for Recommender Systems}. In: {SIGIR}. pp. 770--779 (2022)

\bibitem{Amazon2018}
Ni, J., Li, J., McAuley, J.J.: Justifying recommendations using distantly-labeled reviews and fine-grained aspects. In: {EMNLP-IJCNLP}. pp. 188--197. Association for Computational Linguistics (2019)

\bibitem{DBLP:journals/vldb/PitouraSK22}
Pitoura, E., Stefanidis, K., Koutrika, G.: Fairness in rankings and recommendations: an overview. {VLDB} J.  \textbf{31}(3),  431--458 (2022)

\bibitem{rastegarpanah2019fighting}
Rastegarpanah, B., Gummadi, K.P., Crovella, M.: Fighting fire with fire: Using antidote data to improve polarization and fairness of recommender systems. In: WSDM. pp. 231--239 (2019)

\bibitem{DBLP:conf/sigmod/RoyACDY10}
Roy, S.B., Amer{-}Yahia, S., Chawla, A., Das, G., Yu, C.: Constructing and exploring composite items. In: {SIGMOD}. pp. 843--854 (2010)

\bibitem{MatrixFactAlgo}
Salakhutdinov, R., Mnih, A.: Probabilistic matrix factorization. In: NeurIPS. pp. 1257--1264 (2007)

\bibitem{DBLP:conf/sigmod/SalimiRHS19}
Salimi, B., Rodriguez, L., Howe, B., Suciu, D.: Interventional fairness: Causal database repair for algorithmic fairness. In: {SIGMOD}. pp. 793--810 (2019)

\bibitem{SinghJ18}
Singh, A., Joachims, T.: Fairness of exposure in rankings. In: {SIGKDD}. pp. 2219--2228 (2018)

\bibitem{SunFYFQOL24revisiting}
Sun, Z., Feng, K., Yang, J., Fang, H., Qu, X., Ong, Y., Liu, W.: Revisiting bundle recommendation for intent-aware product bundling. Trans. Recomm. Syst.  \textbf{2}(3),  24:1--24:34 (2024)

\bibitem{XieLW12}
Xie, M., Lakshmanan, L.V.S., Wood, P.T.: Composite recommendations: from items to packages. Frontiers Comput. Sci.  \textbf{6}(3),  264--277 (2012)

\bibitem{yao2017beyond}
Yao, S., Huang, B.: Beyond parity: Fairness objectives for collaborative filtering. NeurIPS  \textbf{30} (2017)

\bibitem{yelpDataset}
Yelp: Dataset. \url{https://business.yelp.com/data/resources/open-dataset/}

\end{thebibliography}


%

\newpage

{\centering \huge
    \textbf{Supplementary Material}
}

\setcounter{section}{0}
\renewcommand*{\thesection}{\Alph{section}}

\section{Experimental details}

\subsection{Description of the datasets} \label{sec:app_datasets}

\subsubsection{MovieLens.}
\begin{figure}[ht]
\centering
\includegraphics[width=0.7\linewidth]{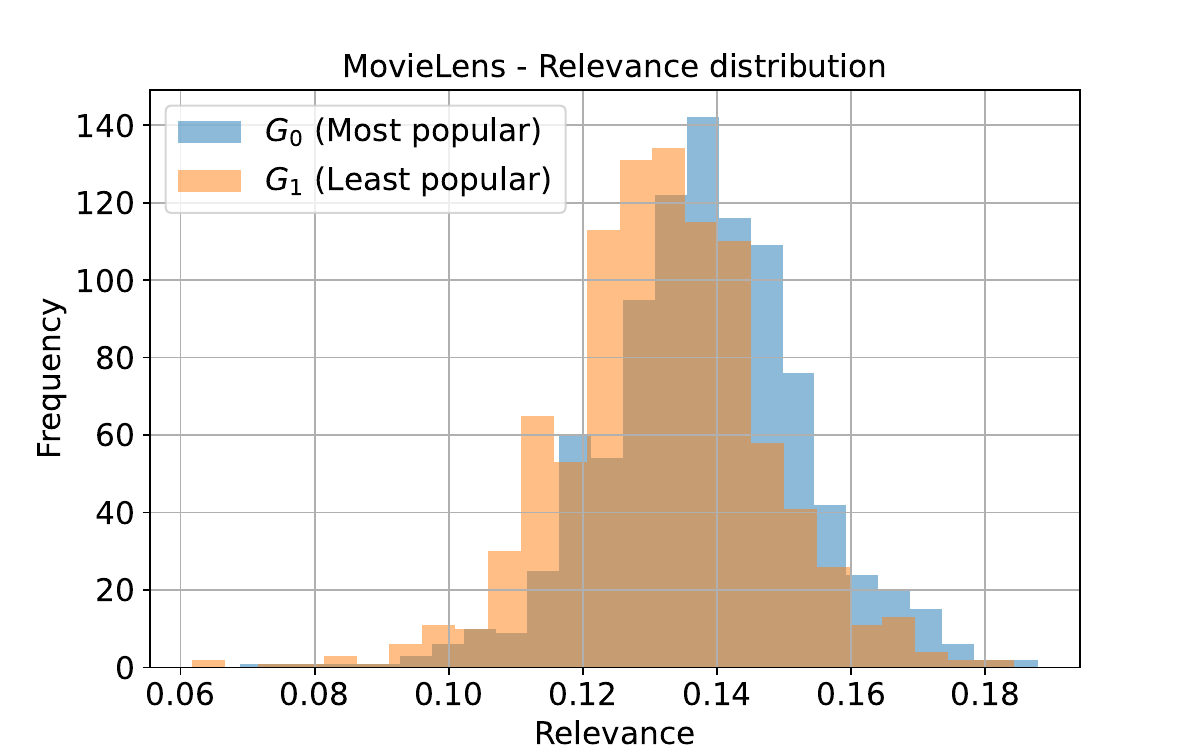}
\caption{Relevance score distributions for \textit{MovieLens}.}
\label{fig:ml_dist_rel}
\end{figure}

\underline{Compatibility:} We use a notion of similarity between movies as compatibility. We consider that two movies are similar if they are from the same period and have similar genres. We compute the similarity between two movies $i$ and $j$ w.r.t. period as $\text{sim}^P(i,j) = \exp(-\vert\text{year}_i - \text{year}_j\vert / 10)$, where $year_i$ is the year of movie $i$. As movies have multiple genres, we one-hot encode the genres of each movie into a vector $genre_i$ and  compute "genre similarity" as the Jaccard similarity between two genre vectors: $\text{sim}^G(i,j) = \text{Jaccard}(\text{genre}_i, \text{genre}_j)$. Finally, we compute the overall similarity as $\text{sim}(i,j) = 0.5 \times \text{sim}^P(i,j) + 0.5 \times \text{sim}^G(i,j)$.
\underline{Fairness:} We consider fairness of exposure with respect to movie popularity: indeed, recommendation systems are known to be prone to popularity bias, that is, to over-recommend the most popular items to the detriment of the least popular ones (or alternatively attributing higher relevance scores to the most popular items). We can see this at play in the \textit{MovieLens} dataset (the relevance scores output by the single item SVD recommender are generally higher for popular movies, as shown in Figure~\ref{fig:ml_dist_rel}). This can create an unwelcome phenomenon of "rich get richer". To build the fairness groups $G_0, G_1$ based on movie popularity, we fetch additional data from the \textit{iMDb} API, in particular the number of users who have rated a movie. Based on preliminary observations, we set a threshold of 20,000 votes: movies with more than 20,000 votes belong to $G_0$, and movies with less than 20,000 votes belong to $G_1$ (the "discriminated" group).

\subsubsection{Yelp.}
\underline{Pre-processing:} We limit the scope to open businesses (restaurants, bars, etc.) in the city of Philadelphia that have received more than 20 reviews. Moreover, we only keep users who have reviewed more than 20 businesses.\\
\underline{Compatibility:} We measure the compatibility between two businesses as the distance between their locations. We compute the distance between two locations from their longitude and latitude using the Haversine formula. Denoting $d(i,j)$ the distance in kilometers between two businesses $i$ and $j$, we compute compatibility as $\exp(-\vert d(i,j)\vert / 3)$ to obtain a compatibility score between 0 and 1.
\underline{Fairness:} As for \textit{MovieLens}, we split the business into different popularity tiers $G_0, G_1, G_2$. We measure popularity based on the number of reviews for a given business: businesses with more than 500 reviews are assigned to $G_0$, businesses between 100 and 500 reviews are assigned to $G_1$, and businesses with less than 100 reviews are assigned to $G_2$.
\underline{Complementarity:} We use the type of businesses from the dataset to get 12 large categories of businesses: \textit{Coffee}, \textit{Bar}, \textit{Brunch}, \textit{Italian}, \textit{Sandwich}, \textit{Meat}, \textit{Seafood}, \textit{Asian}, \textit{TexMex}, \textit{Vegetarian}, \textit{Healthy}, \textit{Sweet}.

\subsubsection{Amazon.}
\underline{Pre-processing:} We limit the scope to items from the category "Cell Phones \& Accessories". We limit the scope to items from the most common brands (those with more than 30 items) from the following countries: USA, China, Japan and South Korea. The majority of items come from these countries.
\underline{Compatibility:} We use the features \textit{also buy} and \textit{also view} to measure compatibility between two items. For a given product $i$, \textit{also buy} is the list of products that have also been bought by users buying $i$, while \textit{also view} is the list of products that have been viewed by users buying $i$. We consider that \textit{also buy} is a stronger relationship between two items than \textit{also view}. To compute compatibility between two products $i$ and $j$, we first compute matrix $s(i,j)$ such that $s(i,j) = 1$ if $j$ is in the \textit{also buy} list of $i$ and $s(i,j) = 0.5$ if $j$ is in the \textit{also view} list of $i$. As the lists are not necessarily symmetric ($j$ can be in the \textit{also buy} list of $i$ but not \textit{vice versa}), we then make the matrix symmetric and normalize it to obtain compatibility scores between 0 and 1.
\underline{Fairness:} We determine the $K=4$ groups based on their country of origin. $G_0$ corresponds to items from Chinese brands, $G_1$ corresponds to items from Japanese brands, $G_2$ corresponds to items from South Korean brands and $G_3$ corresponds to items from American brands. We have $\vert G_0 \vert = 176$, $\vert G_1 \vert = 179$, $\vert G_2 \vert = 142$, $\vert G_3 \vert = 793$: there is a majority of items from American brands (61\%). Our fairness goal for this dataset is therefore to ensure a more balanced representation of items from different origins (for instance, at least $50\%$ of items from China, Japan, and South Korea).
\underline{Complementarity:} We use the product types given in the dataset, \textit{e.g.}, \textit{Cases}, \textit{Batteries}, \textit{Headphones}, \textit{Speakers}. There are $Z=62$ different types of items.

\subsection{Complementarity constraints}

\subsubsection{Yelp.} For \textit{Small} bundles ($L=6$), we use the following $L_z$'s:\\ $(2, 2, 1, 1, 1, 1, 1, 1, 1, 1, 2, 1)$. For \textit{Large} bundles ($L=6$), we use the following $L_z$'s: $(3, 3, 2, 2, 2, 1, 1, 2, 1, 1, 3, 1)$.

\subsubsection{Amazon.} We use $L_z=1$ for each product type, such that a bundle is made of items from strictly different types.
\section{Proofs}\label{sec:proofs}

\begin{proposition}
    Demographic parity (Definition~\ref{def:demographic_parity}) is equivalent to:
    \[
        E_k = p_k \enspace, \enspace k=1,...,K \Longleftrightarrow E_k \ge p_k \enspace, \enspace k=1,...,K\enspace.
    \]
\end{proposition}

\begin{proof}
Indeed, if $E_k = p_k$ for every $k$, then $E_k/p_k=1$ for every $k$, and therefore $ \frac{E_k}{p_k} = \frac{E_\ell}{p_\ell}$ for any pair $(k, \ell)$. Conversely, suppose that we have demographic parity. We have:
\begin{align*}
    \frac{E_k}{p_k} = \frac{1 - \sum_{\ell \ne k} E_\ell}{1 - \sum_{\ell \ne k}p_\ell} &\Longleftrightarrow E_k - E_k\sum_{\ell \ne k}p_\ell = p_k - p_k  \sum_{\ell \ne k} E_\ell \\
    &\Longleftrightarrow E_k = p_k + E_k\sum_{\ell \ne k}p_\ell - p_k  \sum_{\ell \ne k} E_\ell \enspace.
\end{align*}
Now we show that $E_k\sum_{\ell \ne k}p_\ell - p_k  \sum_{\ell \ne k} E_\ell = 0$. We have:
\begin{align*}
    E_k\sum_{\ell \ne k}p_\ell - p_k  \sum_{\ell \ne k} E_\ell &= E_k p_k \left(\sum_{\ell \ne k}\frac{p_\ell}{p_k} - \sum_{\ell \ne k} \frac{E_\ell}{E_k} \right) \\
    &= E_k p_k \left(\sum_{\ell \ne k}\frac{p_\ell}{p_k} - \frac{E_\ell}{E_k} \right) \enspace.
\end{align*}
But, from demographic parity, $\frac{E_\ell}{E_k} = \frac{p_\ell}{p_k}$ and therefore $\sum_{\ell \ne k}\frac{p_\ell}{p_k} - \frac{E_\ell}{E_k} = 0$, which yields the first equivalence.

Now, we show that:
\[
    E_k = p_k \enspace, \enspace k=1,...,K \Longleftrightarrow E_k \ge p_k \enspace, \enspace k=1,...,K \enspace.
\]
We just have to show that $\{E_k \ge p_k \enspace, \enspace k=1,...,K\} \implies  \{E_k = p_k \enspace, \enspace k=1,...,K \}$, as the other implication is obvious. 

We have $E_k = 1 - \sum_{\ell \ne k} E_\ell$. But as $E_\ell \ge p_\ell$ for every $\ell \ne k$, then $\sum_{\ell \ne k} E_\ell \ge \sum_{\ell \ne k} p_\ell$, and thus $E_k \le 1 - \sum_{\ell \ne k} p_\ell = p_k$. Therefore, for any $k$, we have both $E_k \ge p_k$ and $E_k \le p_k$, that is $E_k = p_k$.

\end{proof}

\begin{proposition}
    The online constraint~(\ref{eq:const_cond_pfair}) can be derived from the global constraint~(\ref{eq:const_global_pfair}). As a result, meeting constraint~(\ref{eq:const_cond_pfair}) for every user $u=1,...,T$ ensures that constraint~(\ref{eq:const_global_pfair}) is met.
\end{proposition}

\begin{proof}
    Let $N(u) = \vert B_u\vert$ and $N_k(u) = \sum_{i \in B_u} g_{ik}$ denote respectively the total number of items in $B_u$ and number of items from group $G_k$ in $B_u$, and $E_k(u) = \frac{N_k(u)}{\vert B_u\vert}$ the exposure of group $G_k$ in $B_u$. Let us further define $N_k(1\!:\!u-1) = \sum_{t=1}^{u-1} N_k(t)$, $N(1\!:\!u) = \sum_{t=1}^{u-1} N(t)$, and $E_k(1\!:\!u-1) = \frac{N_k(1\!:\!u-1)}{N(1\!:\!u-1)}$. 
    
    We have:
\begin{align*}
    E_k(1\!:\!u)& = \frac{N(1\!:\!u-1)}{N(1\!:\!u)} E_k(1\!:\!u-1) + \frac{\vert B_u \vert}{N(1\!:\!u)} E_k(u) \\
    &= \frac{N_k(1\!:\!u-1)}{N(1\!:\!u)} + \frac{N_k(u)}{N(1\!:\!u)} \enspace.
\end{align*}
Thus, if we want the recommendations to comply with the P-fairness constraint at step $u$, we want $E_k(1\!:\!u) \ge \rho_k(1-\epsilon)$, which gives:
\begin{align*}
    \quad N_k(1\!:\!u-1) + N_k(u) &\ge N(1\!:\!u)\rho_k(1-\epsilon) \\ 
    \Longleftrightarrow N_k(1\!:\!u-1) + N_k(u) &\ge N(1\!:\!u-1)\rho_k(1-\epsilon) + \vert B_u\vert \rho_k(1 -  \epsilon) \\
    \Longleftrightarrow N_k(u) - \vert B_u \vert \rho_k(1 - \epsilon) &\ge N(1\!:\!u-1)\rho_k(1-\epsilon) - N_k(1\!:\!u-1) \enspace.
\end{align*}
Denoting $\Delta_k(u) = N_k(u) - \vert B_u \vert \rho_k(1 - \epsilon)$ and $\Delta_k(1\!:\!u-1) = N_k(1\!:\!u-1) - N(1\!:\!u-1)\rho_k(1-\epsilon)$, we can rewrite the P-fairness constraint as: 
\begin{align*}
\Delta_k(u) + \Delta_k(1\!:\!u-1) \ge 0. 
\end{align*}

Moreover, if we meet the constraint $E_k(1\!:\!u) \ge \rho_k (1 - \epsilon)$ for each user $u$, therefore in particular $E_k = E_k(1\!:\!T) \ge \rho_k(1-\epsilon)$.
\end{proof}
\newpage
\section{Additional Figures and Tables} \label{sec:app_figs_tabs}

\begin{figure}[ht]
\centering
\begin{minipage}{.48\textwidth}
\includegraphics[width=1.05\linewidth]{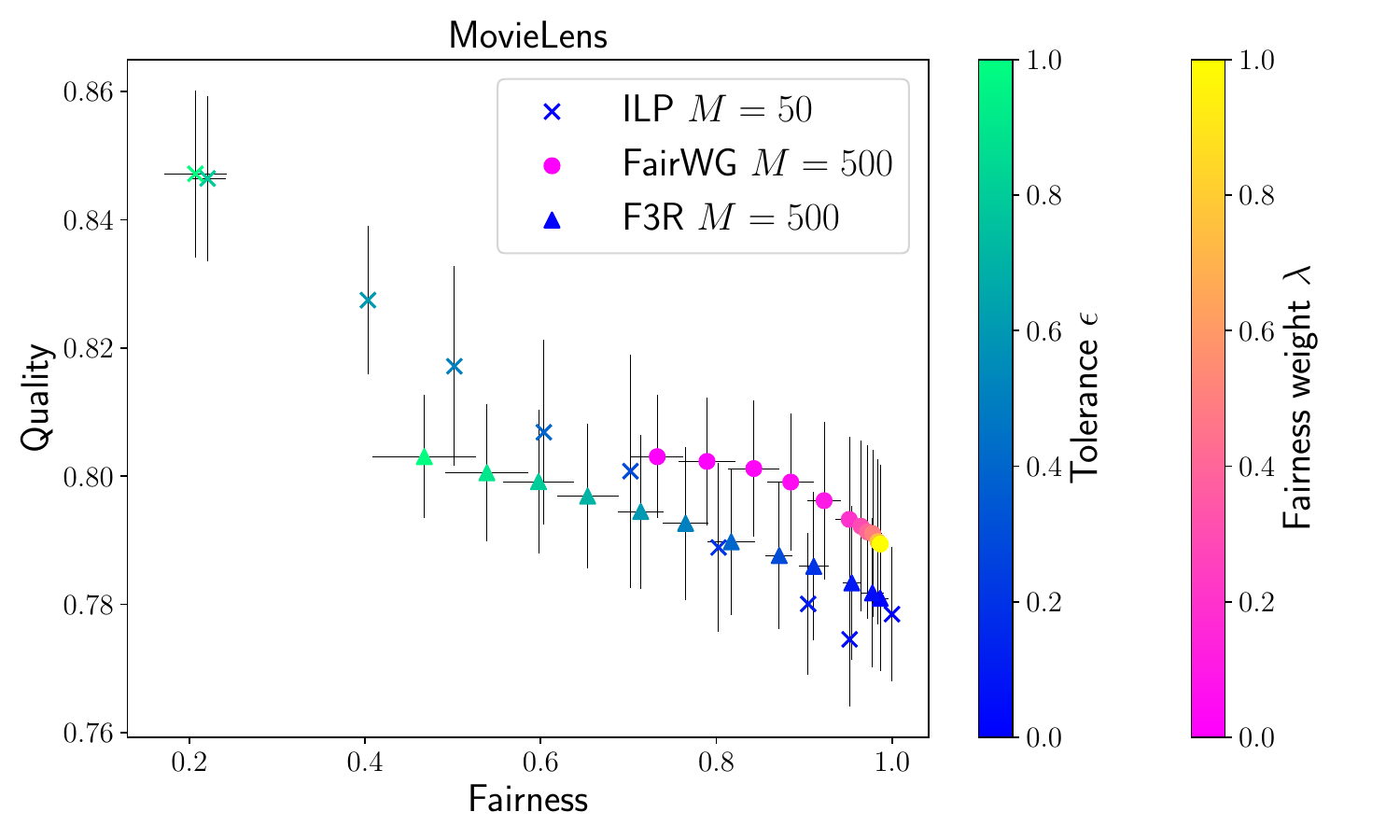}
\end{minipage} \hspace{0.2cm}
\begin{minipage}{.48\textwidth}
\includegraphics[width=1.05\linewidth]{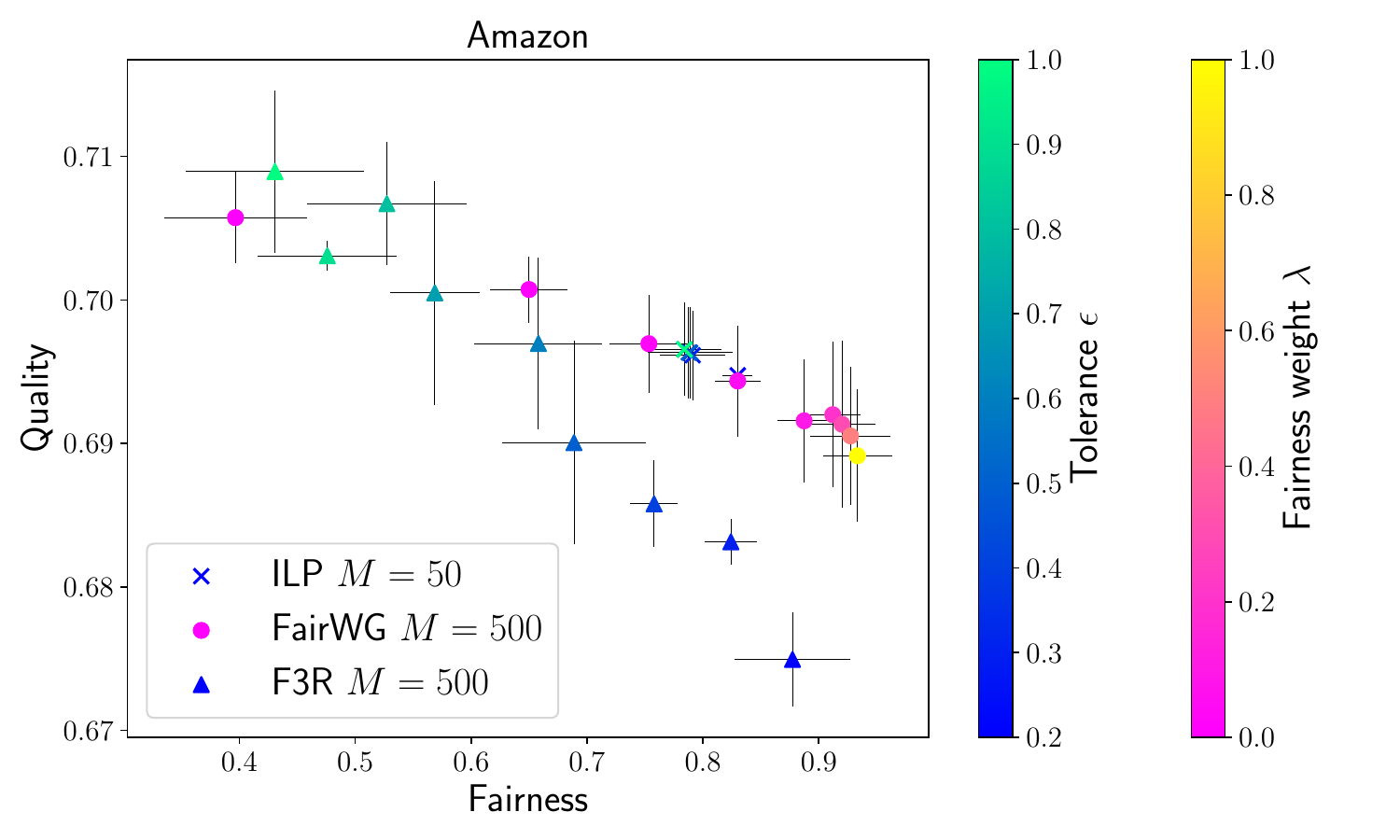}
\end{minipage}
\caption{Fairness-quality trade-offs on \textit{MovieLens} and \textit{Amazon} for comparable runtimes ($\approx1s$/user for \textsc{ILP} with $M=50$, $\lessapprox 10^{-2}s$/user for \textsc{FairWG} and \textsc{F3R} with $M=500$.)}
\label{fig:ilp_heuristic_comp_sameruntime_other_dataset}
\end{figure}

\begin{table*}[ht]
\begin{center}
\tiny
\begin{tabular}{@{}l@{\hspace{20pt}}c@{\hspace{10pt}}c@{\hspace{12pt}}c@{\hspace{20pt}}c@{\hspace{12pt}}c@{\hspace{20pt}}c@{}}
\toprule
\multicolumn{7}{c}{\textbf{ILP}} 
\\ \midrule
              & \multicolumn{2}{c}{MovieLens}                   & \multicolumn{2}{c}{Yelp}             & \multicolumn{2}{c}{Amazon}           \\ \midrule
  $M$ &\textit{ Runtime (s)} &  $\Delta$ Quality & Runtime (s) & $\Delta$ Quality & Runtime (s) &  $\Delta$ Quality  \\ \midrule
$50$ & $0.68 \pm 0.29$ & --- & $1.03 \pm 0.37$ &   ---   &      $0.14 \pm 0.04$     &          ---       \\
$100$    & $3.39 \pm2.74$  & $\nearrow 10.53 \% \pm 3.82\%$ & $12.85 \pm 3.80$ &      $\nearrow 3.07 \% \pm 3.40\%$      &     $0.46 \pm 0.11$       &        $\nearrow 5.35 \% \pm 3.52\%$           \\
$200$    & $88.61 \pm 23.24$ & $\nearrow 14.04 \% \pm 5.12\%$ & $184.03 \pm 126.99$ &      $\nearrow 6.15 \% \pm 2.24\%$      &     $2.67 \pm 0.44$       &           $\nearrow 9.44 \% \pm 4.67\%$           \\ \midrule
\multicolumn{7}{c}{\textbf{FairWG}} 
\\ \midrule
              & \multicolumn{2}{c}{MovieLens}                   & \multicolumn{2}{c}{Yelp}             & \multicolumn{2}{c}{Amazon}           \\ \midrule
  $M$ &\textit{ Runtime (s)} &  $\Delta$ Quality & Runtime (s) &  $\Delta$ Quality & Runtime (s) &  $\Delta$ Quality  \\ \midrule
$50$ & \multirow{3}{*}{$<10^{-2}$} & --- &         \multirow{3}{*}{$<10^{-2}$}          &      ---      &      \multirow{3}{*}{$<10^{-2}$}      &     ---     \\
$100$    &   & $\nearrow2.56\% \pm 4.87\%$ &  &  $\nearrow 2.36 \% \pm 4.17\%$ &            &             $\nearrow 0.73 \% \pm 2.71 \%$      \\
$200$    & & $\nearrow4.79\% \pm 5.18\%$ &  & $\nearrow 3.75 \% \pm 4.62\%$  &            &                $\nearrow 2.36 \% \pm 4.16 \%$      \\ \midrule
\multicolumn{7}{c}{\textbf{F3R}}
\\ \midrule
              & \multicolumn{2}{c}{MovieLens}                   & \multicolumn{2}{c}{Yelp}             & \multicolumn{2}{c}{Amazon}           \\ \midrule
  $M$ &\textit{ Runtime (s)} &  $\Delta$ Quality & Runtime (s) &  $\Delta$ Quality & Runtime (s) &  $\Delta$ Quality  \\ \midrule
$50$ & \multirow{3}{*}{$<10^{-2}$} & --- &            \multirow{3}{*}{$<10^{-2}$}       &    ---        &      \multirow{3}{*}{$<10^{-2}$}      &   ---    \\
$100$    &   & $\nearrow 1.28 \% \pm 1.92\%$ &  &     $\nearrow 2.95 \% \pm 2.85 \%$       &            &          $\nearrow 0.60 \% \pm 3.68 \%$         \\
$200$    &  & $\nearrow 2.62 \% \pm 2.03\%$ &  &  $\nearrow 4.85 \% \pm 3.00 \%$ &            &                $\nearrow 2.95 \% \pm 4.45 \%$      \\ \bottomrule
\end{tabular}
\end{center}
\caption{ILP runtime (per user) and bundle quality for different values of $M$.}
\label{tab:ilp_runtime}
\end{table*}


\begin{table*}[ht]
\begin{center}
\begin{tabular}{@{}l@{\hspace{20pt}}c@{\hspace{10pt}}c@{\hspace{10pt}}c@{\hspace{20pt}}c@{\hspace{10pt}}c@{\hspace{10pt}}c@{\hspace{20pt}}c@{\hspace{10pt}}c@{\hspace{10pt}}c@{}}
\toprule
              & \multicolumn{3}{c}{\textit{MovieLens}}           \\ \midrule
       $\Rho$     & $(0.5,0.5)$        & $(0.3, 0.7)$        & $(0.2, 0.8)$    \\ \midrule
$\alpha=\infty$ & $0.84 \pm 0.05$ & $0.80 \pm 0.12$ &   $0.00 \pm 0.00$     \\
$\alpha=2$    & $0.84 \pm0.11$  & $0.80 \pm 0.11$ & $\mathbf{0.79 \pm 0.12}$   \\
\midrule
              &    \multicolumn{3}{c}{\textit{Yelp}}               \\ \midrule
     $\Rho$       & $(1/3, 1/3, 1/3)$ & $(0.25, 0.35, 0.4)$ & $(0.2, 0.3, 0.5)$ \\ \midrule
$\alpha=\infty$    &     $0.88 \pm 0.19$       &     $0.00 \pm 0.00$       &  $0.00 \pm 0.00$            \\
$\alpha=2$   &   $\mathbf{0.91 \pm 0.13}$      &    $\mathbf{0.89 \pm 0.16}$        &       $\mathbf{0.89 \pm 0.15}$        \\
\midrule
         & \multicolumn{3}{c}{\textit{Amazon}}           \\ \midrule
       $\Rho$    & $(0.25,0.25,0.25,0.25)$ & $(0.25,0.25,0.2,0.3)$ & $(0.2,0.2,0.1,0.5)$ \\ \midrule
$\alpha=\infty$ &  $0.00 \pm 0.00$ &    $0.00 \pm 0.00$        &     $0.72 \pm 0.03$       \\
$\alpha=2$     &  $\mathbf{0.72 \pm 0.03}$ &      $\mathbf{0.72 \pm 0.03}$      &       $0.72 \pm 0.04$     \\
\bottomrule
\end{tabular}
\end{center}
\caption{ILP bundle quality on different problem instances, for different values of $\alpha$ and different exposure vectors $\Rho$. The tolerance is fixed to $\epsilon=0.1$ (except for \textit{Amazon} for which it is $\epsilon=0.2$, since no solution is found even with $\alpha=2$ for $\epsilon=0.1$).}
\label{tab:ilp_alpha}
\end{table*}

\begin{figure}[ht]
\centering
\begin{minipage}{.49\textwidth}
\includegraphics[width=\linewidth]{plots/ml/f3r_quality_0.pdf}
\end{minipage}
\begin{minipage}{.49\textwidth}
\includegraphics[width=\linewidth]{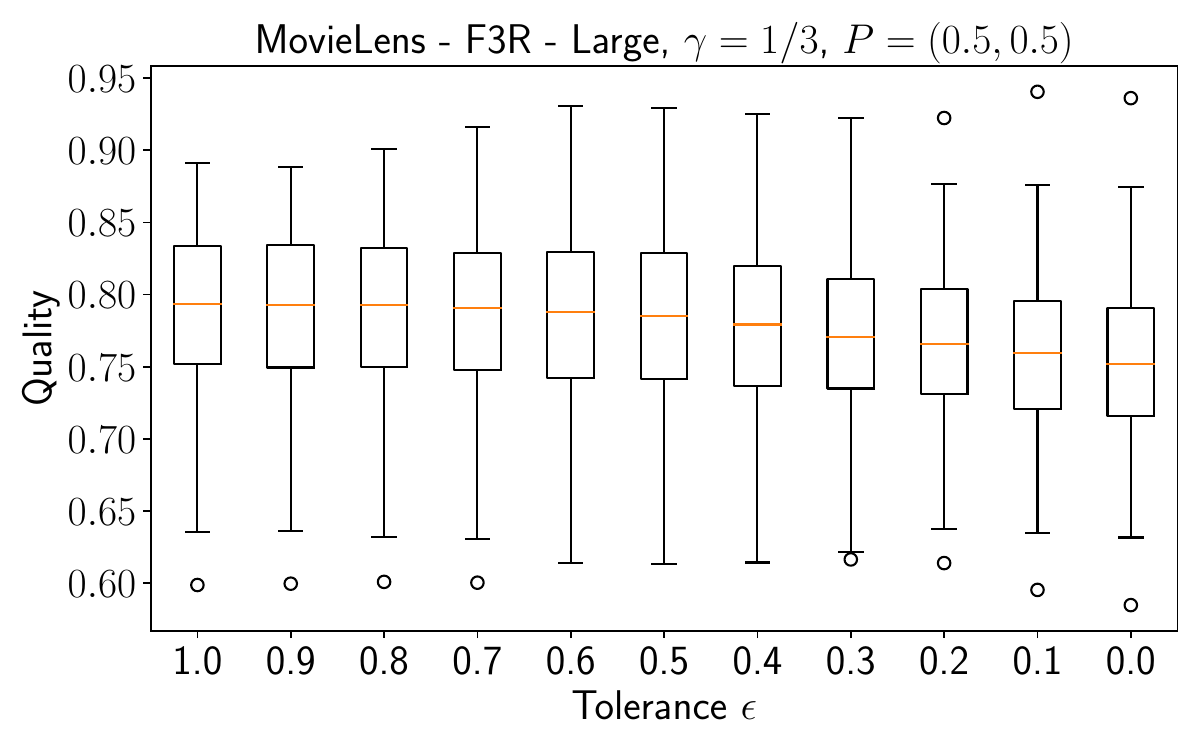}
\end{minipage}
\begin{minipage}{.49\textwidth}
\includegraphics[width=\linewidth]{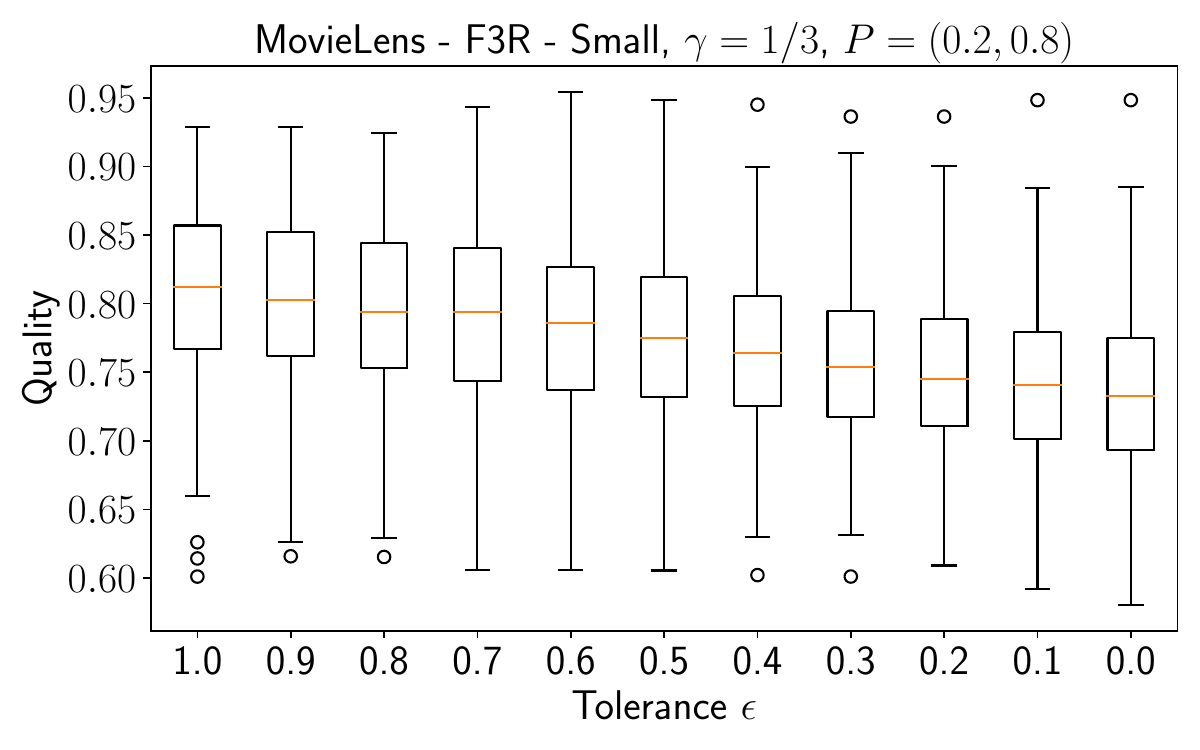}
\end{minipage}
\begin{minipage}{.49\textwidth}
\includegraphics[width=\linewidth]{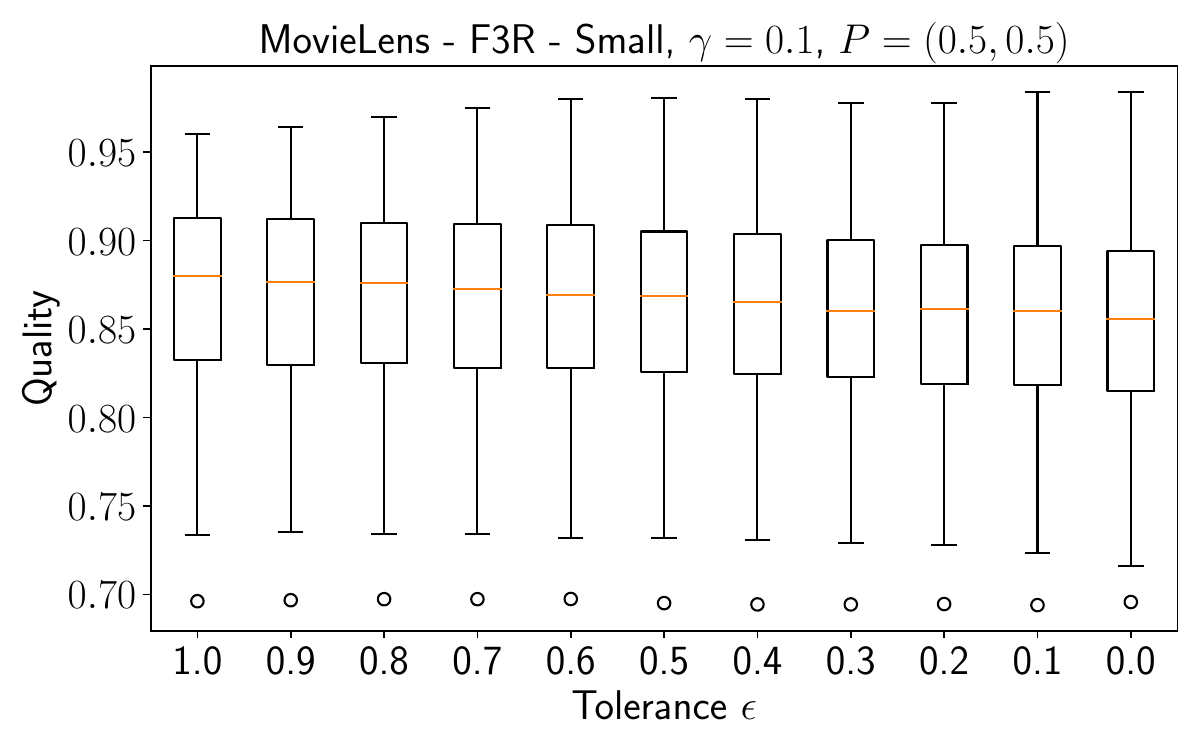}
\end{minipage}
\caption{\textsc{F3R} - Fairness and quality on \textit{MovieLens}.}
\label{fig:app_ml_f3r_quality}
\end{figure}

\begin{figure}[ht]
\centering
\begin{minipage}{\textwidth}
\includegraphics[width=\linewidth]{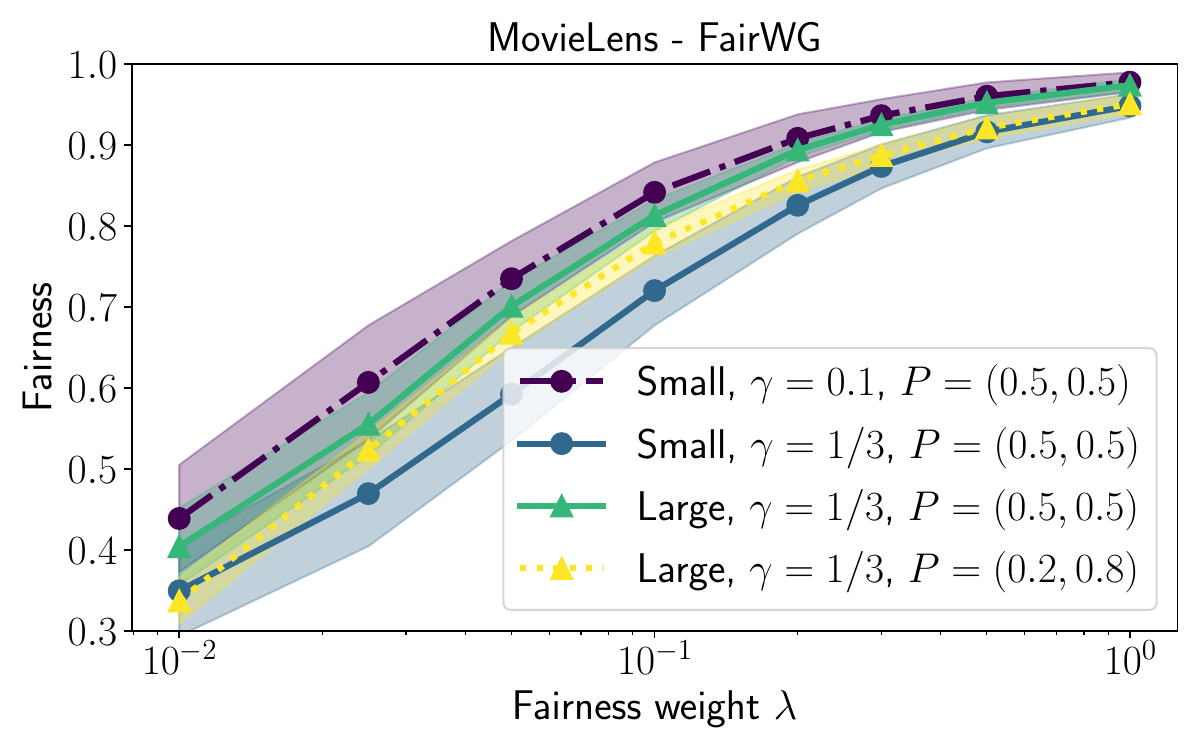}
\end{minipage}
\begin{minipage}{.49\textwidth}
\includegraphics[width=\linewidth]{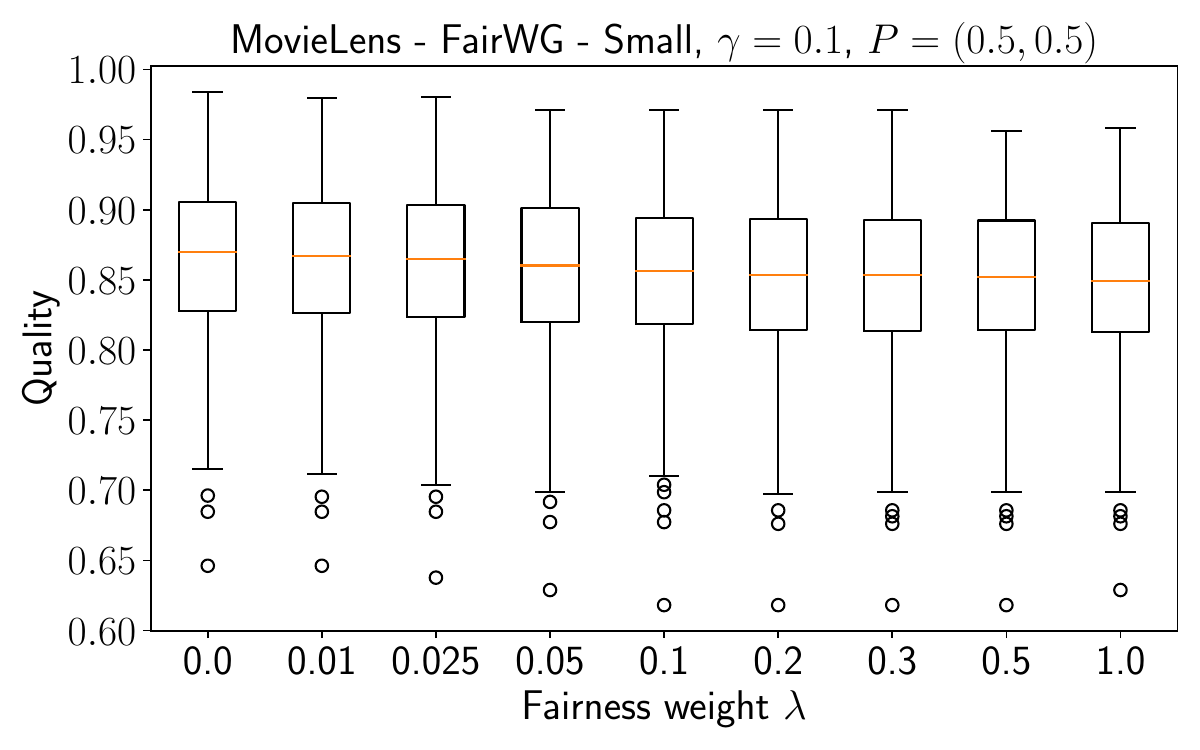}
\end{minipage}
\begin{minipage}{.49\textwidth}
\includegraphics[width=\linewidth]{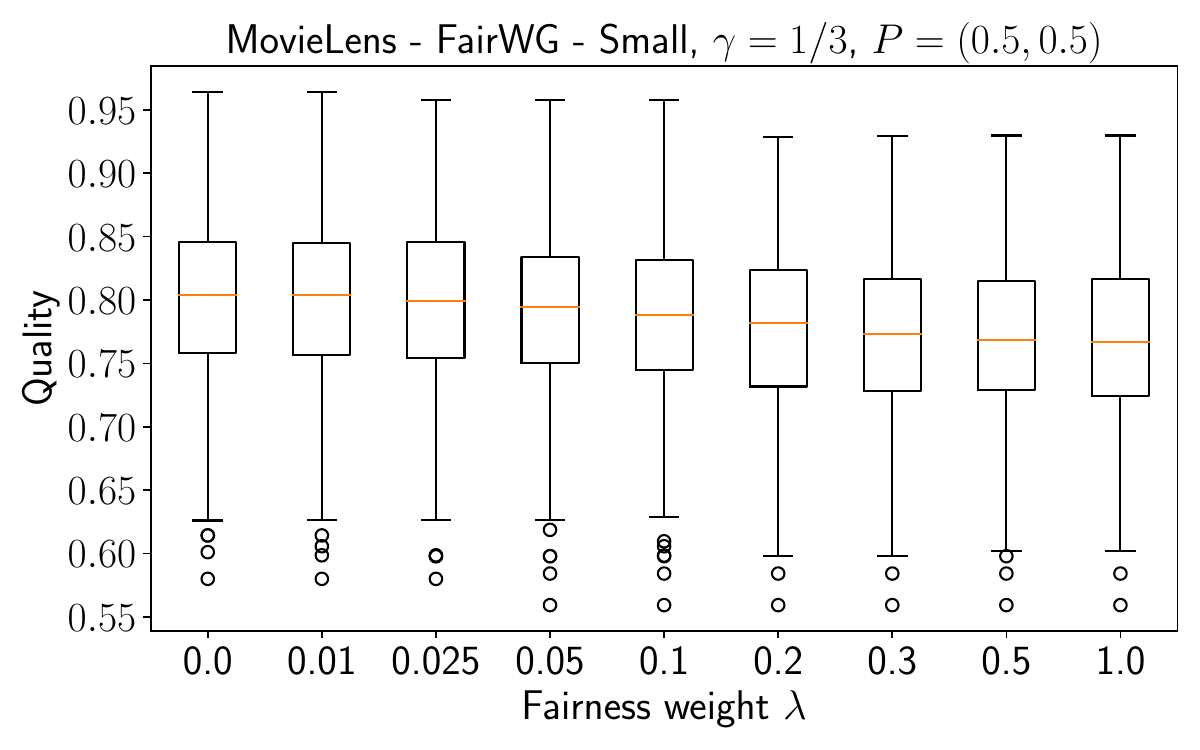}
\end{minipage}
\begin{minipage}{.49\textwidth}
\includegraphics[width=\linewidth]{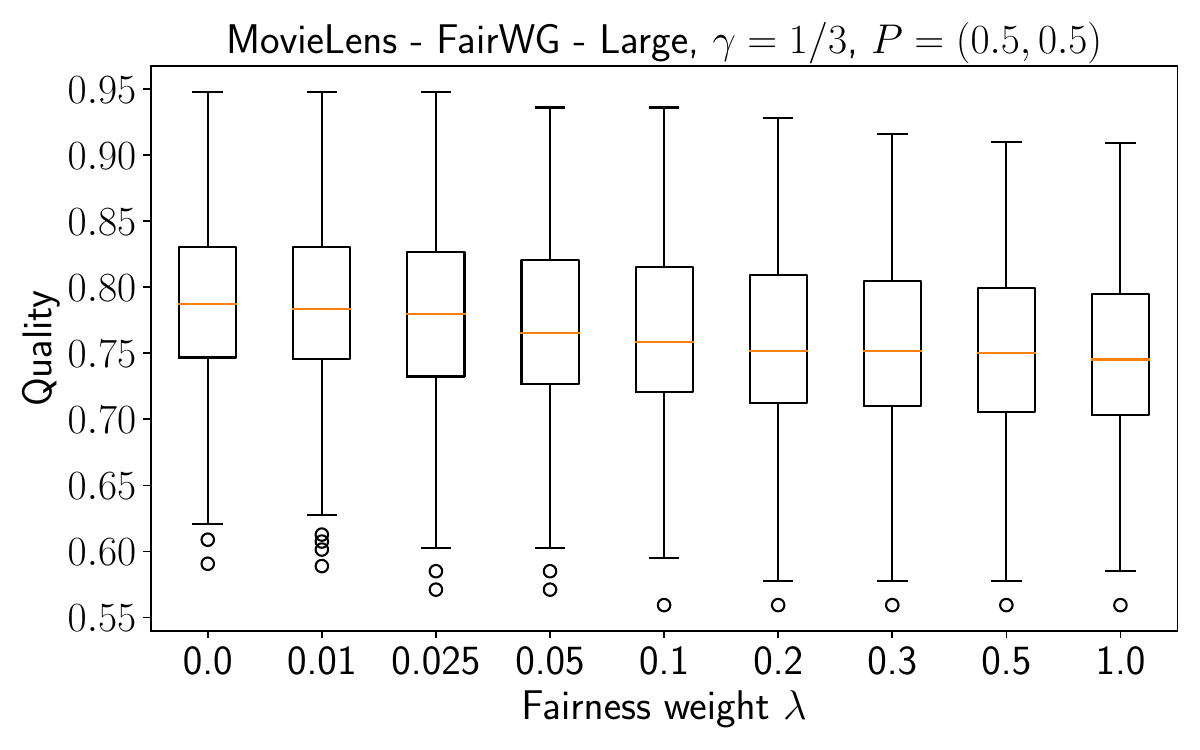}
\end{minipage}
\begin{minipage}{.49\textwidth}
\includegraphics[width=\linewidth]{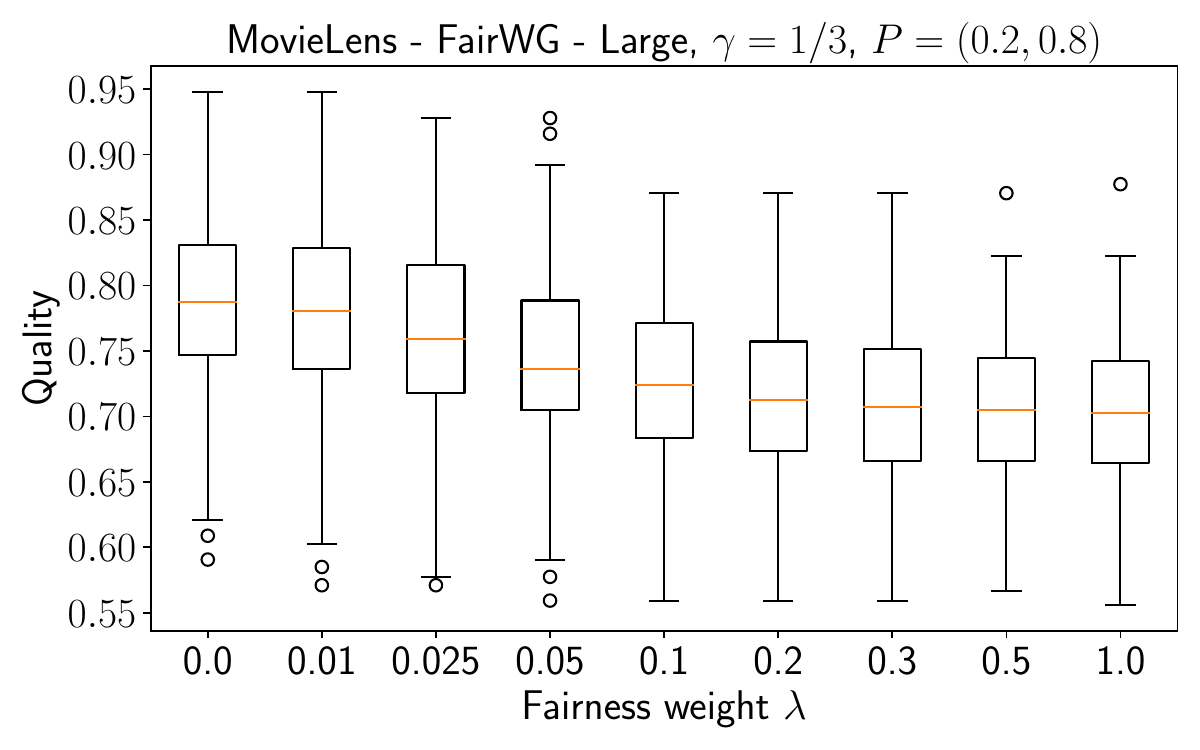}
\end{minipage}
\caption{\textsc{FairWG} - Fairness and quality on \textit{MovieLens}.}
\label{fig:app_ml_fbp_quality}
\end{figure}

\begin{figure}[ht]
\centering
\begin{minipage}{.49\textwidth}
\includegraphics[width=\linewidth]{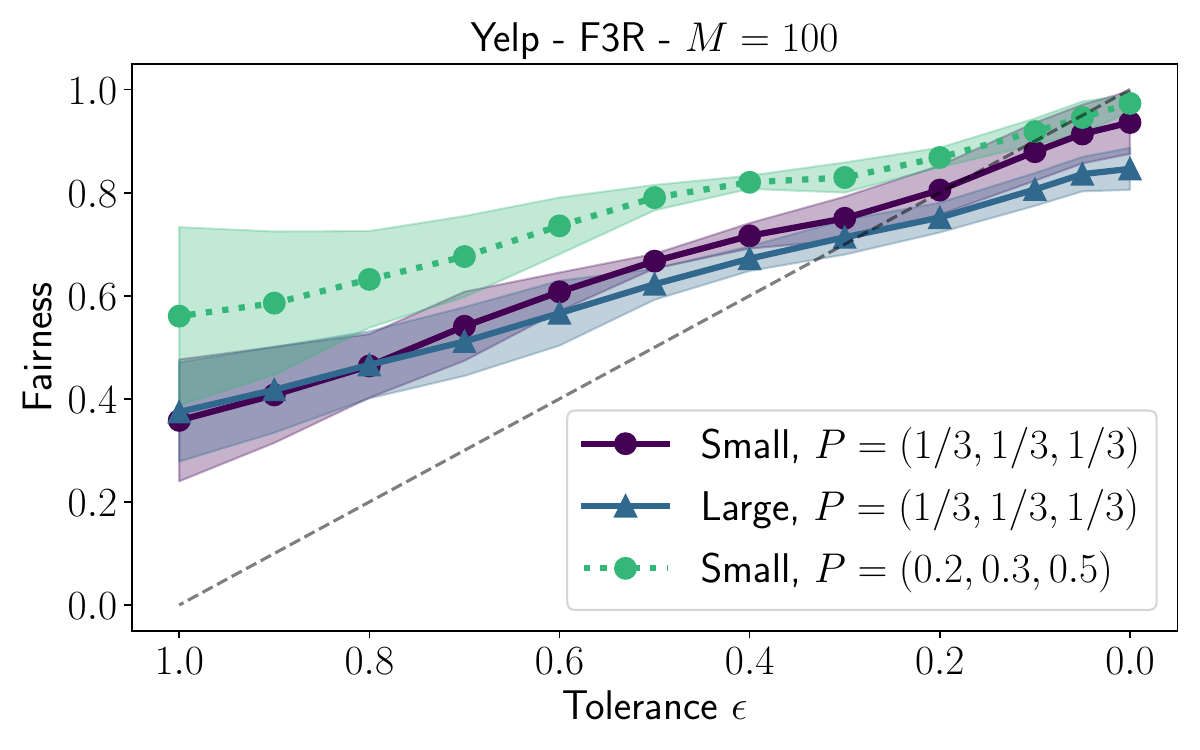}
\end{minipage}
\begin{minipage}{.49\textwidth}
\includegraphics[width=\linewidth]{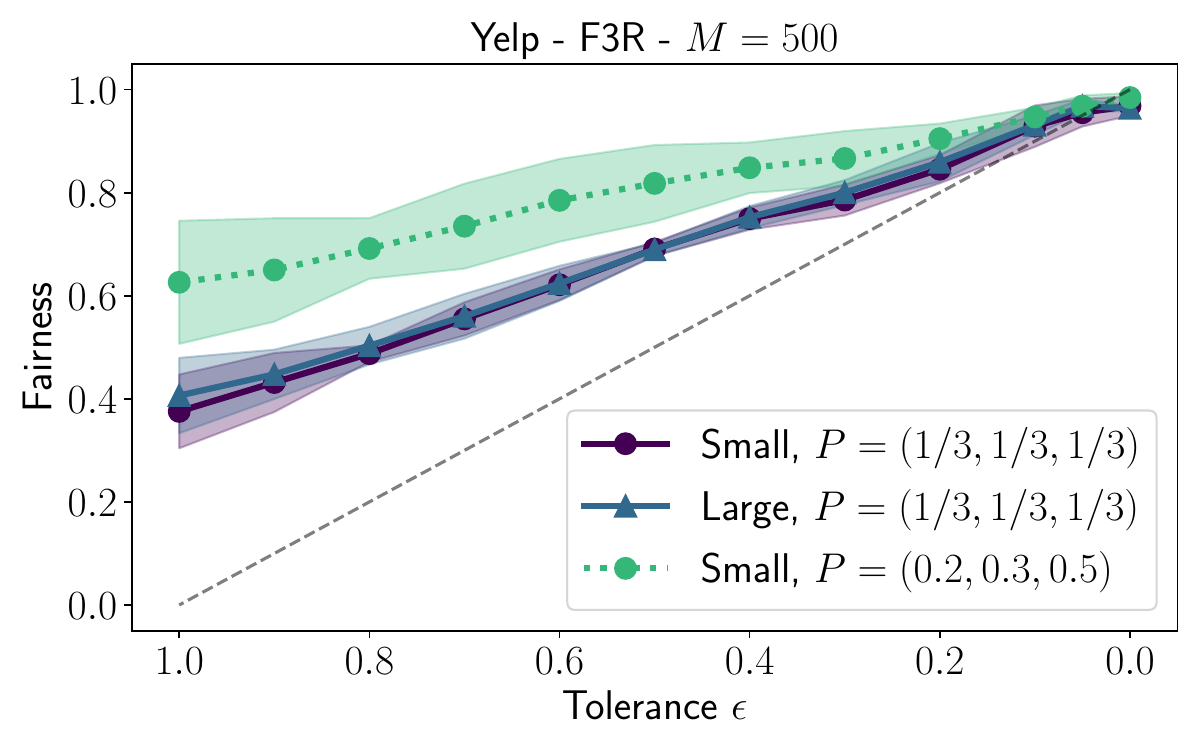}
\end{minipage}
\begin{minipage}{.49\textwidth}
\includegraphics[width=\linewidth]{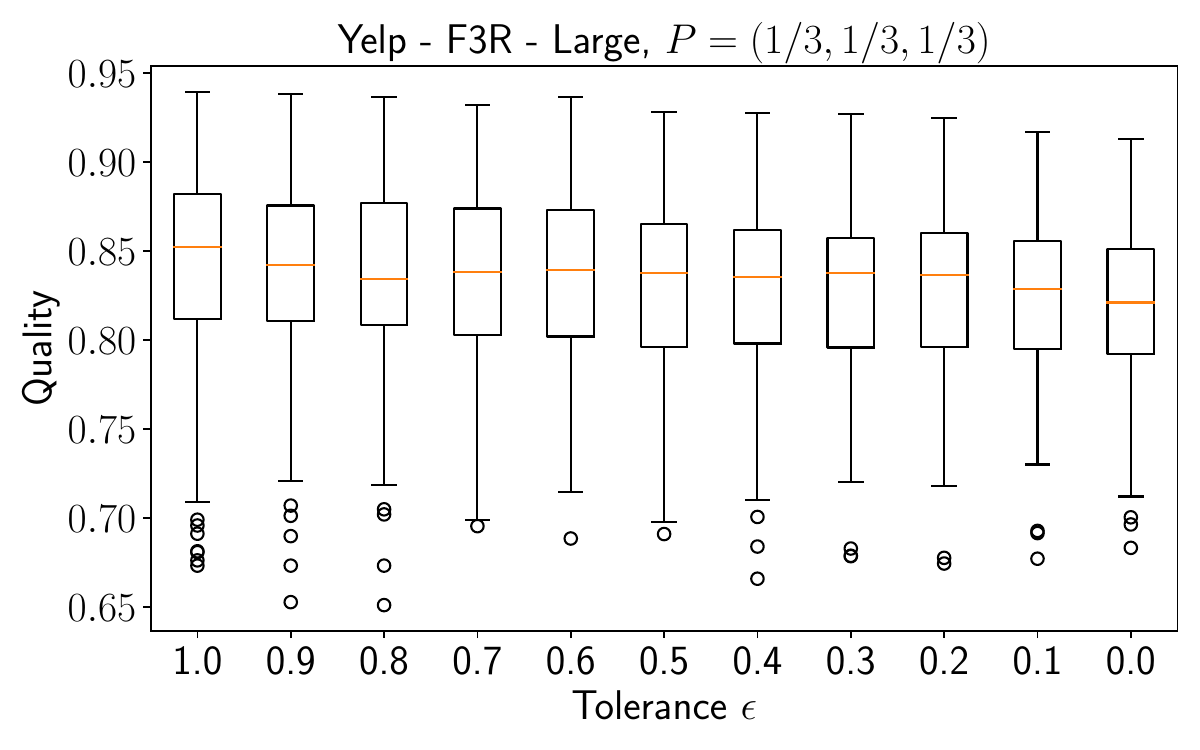}
\end{minipage}
\begin{minipage}{.49\textwidth}
\includegraphics[width=\linewidth]{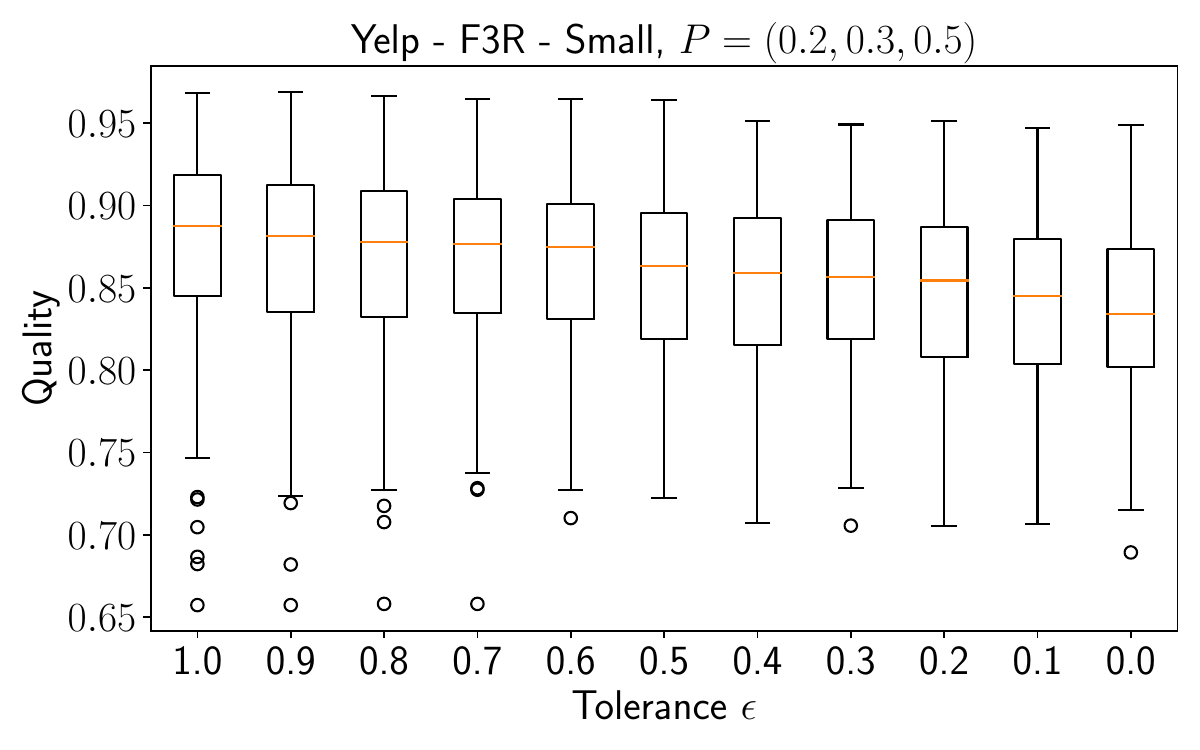}
\end{minipage}
\caption{\textsc{F3R} - Fairness and quality on \textit{Yelp}.}
\label{fig:f3r_yelp}
\end{figure}

\begin{figure}[ht]
\centering
\begin{minipage}{.49\textwidth}
\includegraphics[width=\linewidth]{plots/yelp/fbp_fairness.pdf}
\end{minipage}
\begin{minipage}{.49\textwidth}
\includegraphics[width=\linewidth]{plots/yelp/fbp_quality_0.pdf}
\end{minipage}
\begin{minipage}{.49\textwidth}
\includegraphics[width=\linewidth]{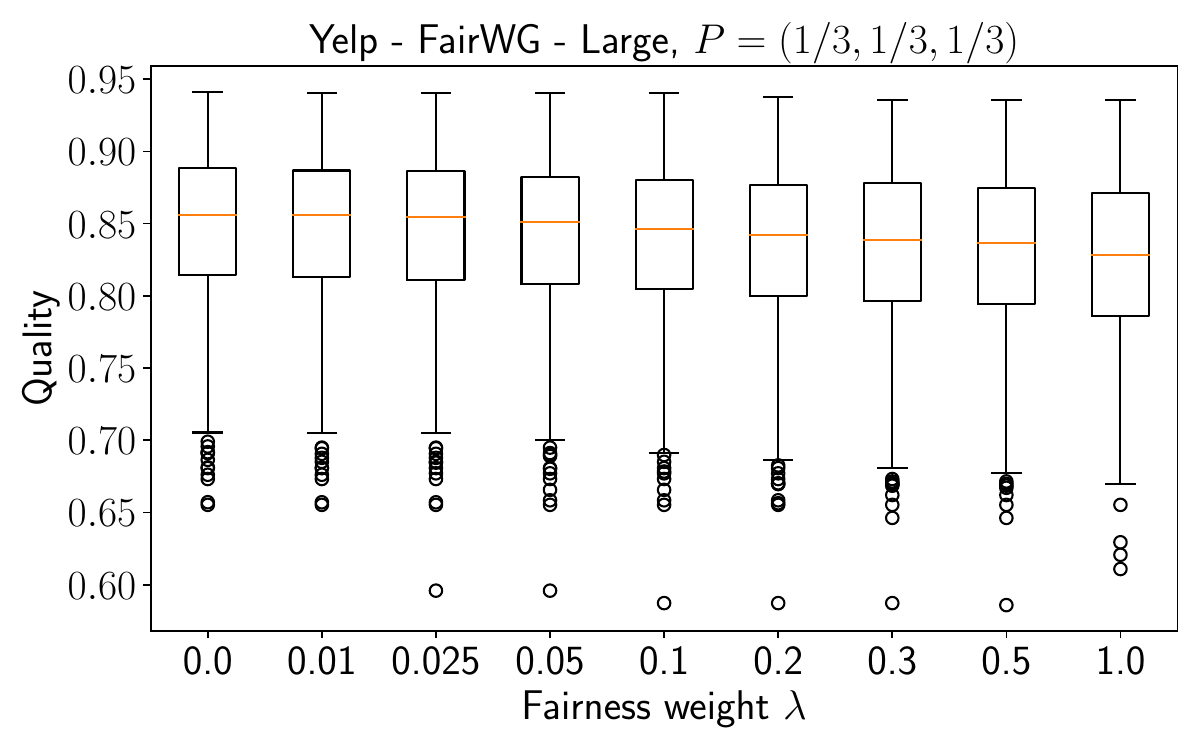}
\end{minipage}
\begin{minipage}{.49\textwidth}
\includegraphics[width=\linewidth]{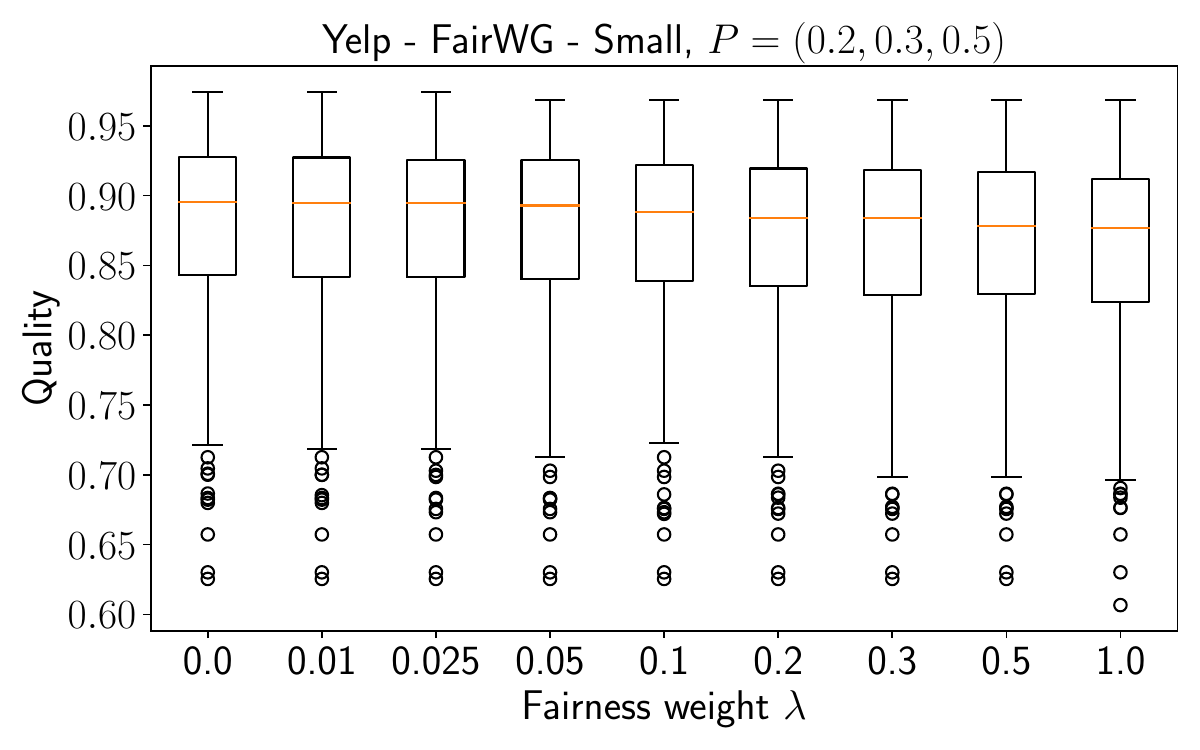}
\end{minipage}
\caption{\textsc{FairWG} - Fairness and quality on \textit{Yelp}.}
\label{fig:fbp_yelp}
\end{figure}

\begin{figure}[ht]
\centering
\begin{minipage}{.49\textwidth}
\includegraphics[width=\linewidth]{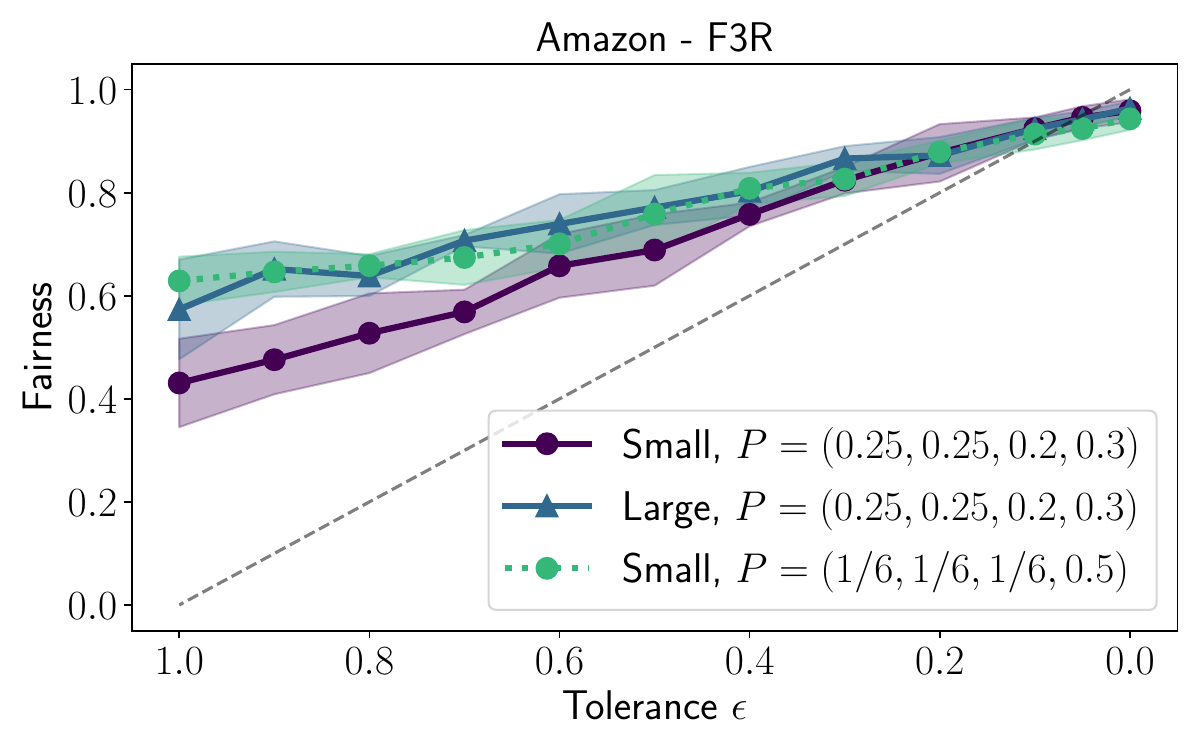}
\end{minipage}
\begin{minipage}{.49\textwidth}
\includegraphics[width=\linewidth]{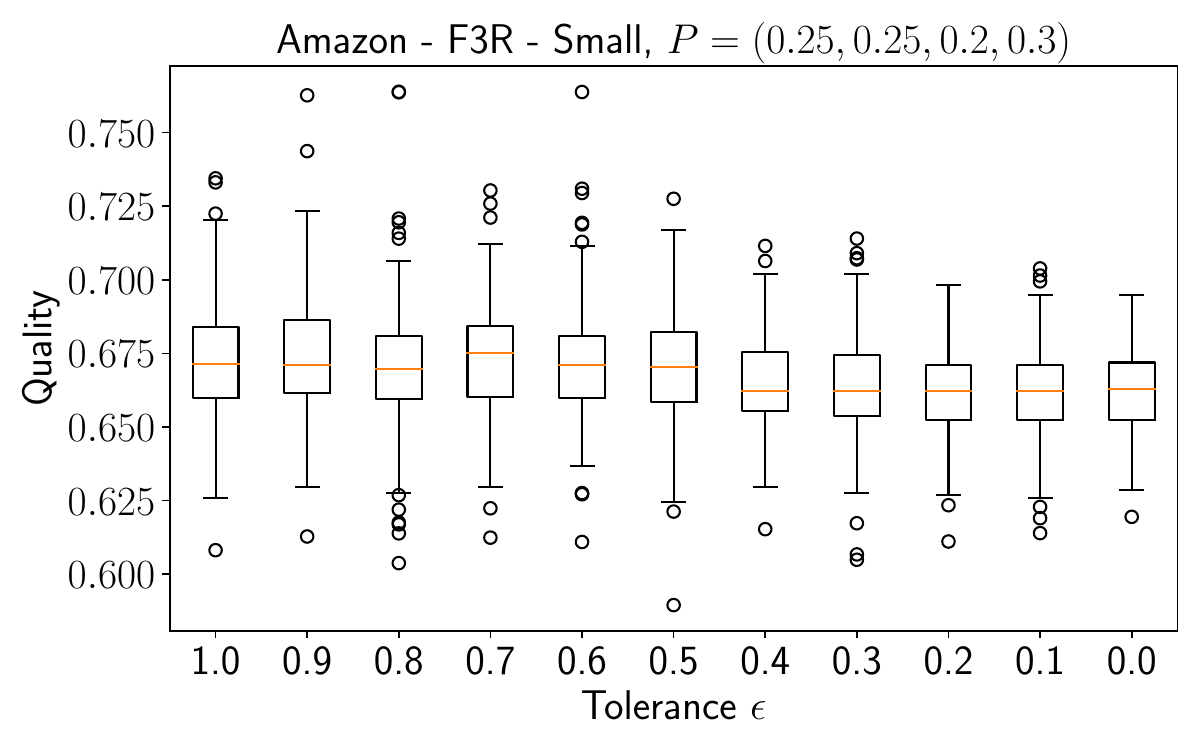}
\end{minipage}
\begin{minipage}{.49\textwidth}
\includegraphics[width=\linewidth]{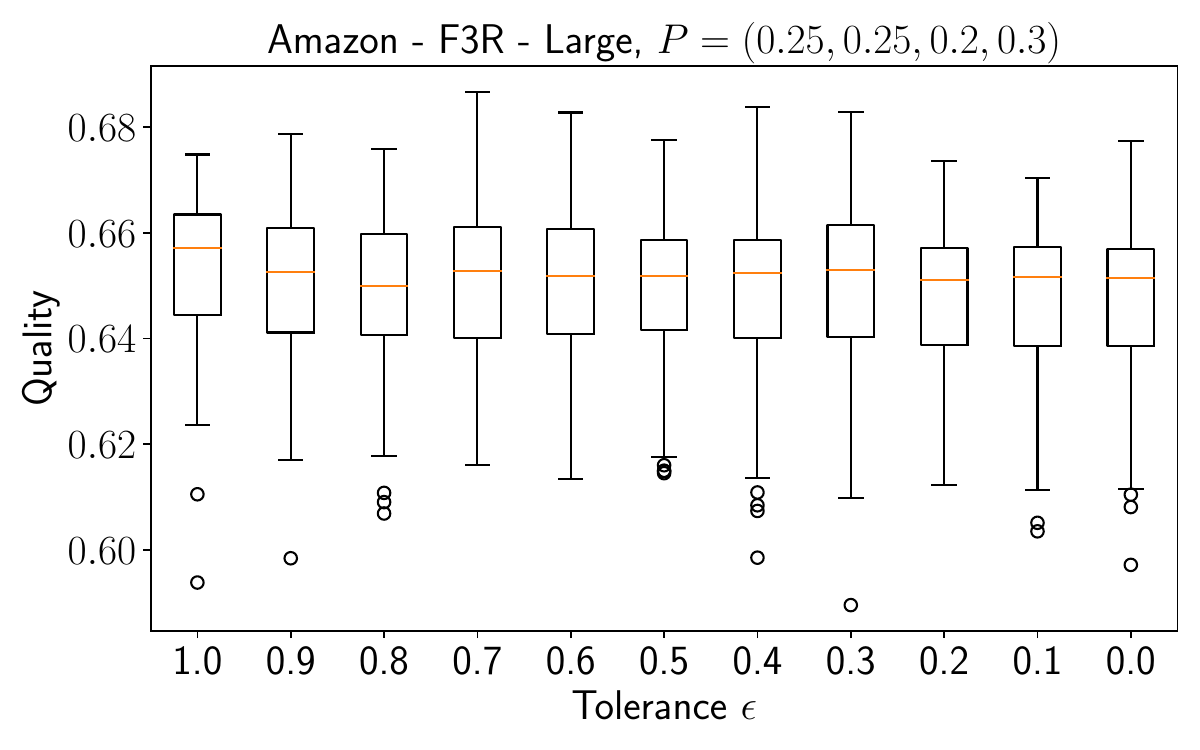}
\end{minipage}
\begin{minipage}{.49\textwidth}
\includegraphics[width=\linewidth]{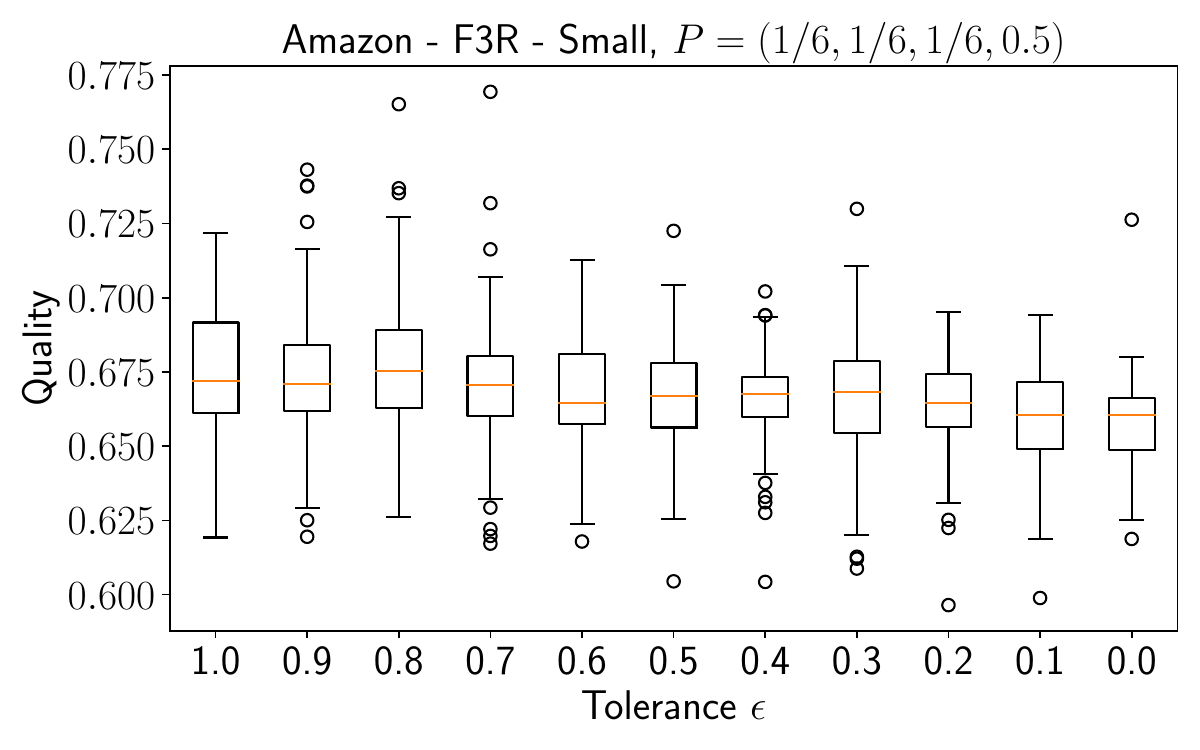}
\end{minipage}
\caption{\textsc{F3R} - Fairness and quality on \textit{Amazon}.}
\label{fig:f3r_amazon}
\end{figure}

\begin{figure}[ht]
\centering
\begin{minipage}{.49\textwidth}
\includegraphics[width=\linewidth]{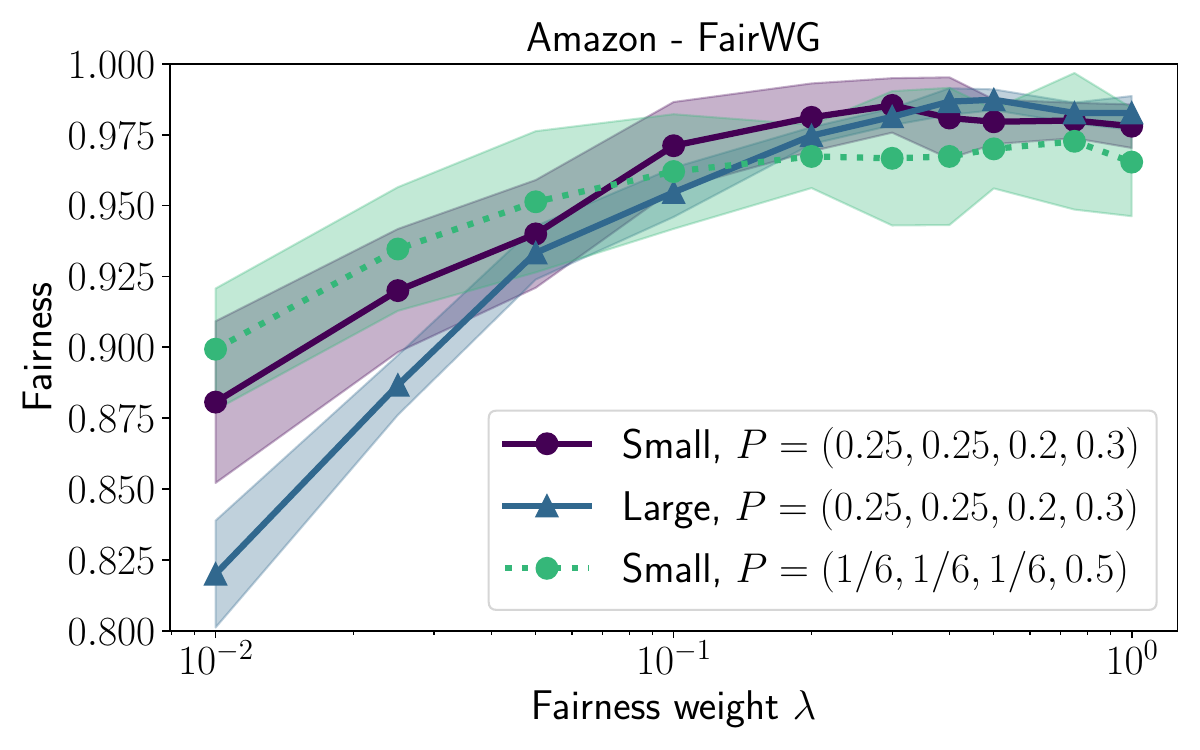}
\end{minipage}
\begin{minipage}{.49\textwidth}
\includegraphics[width=\linewidth]{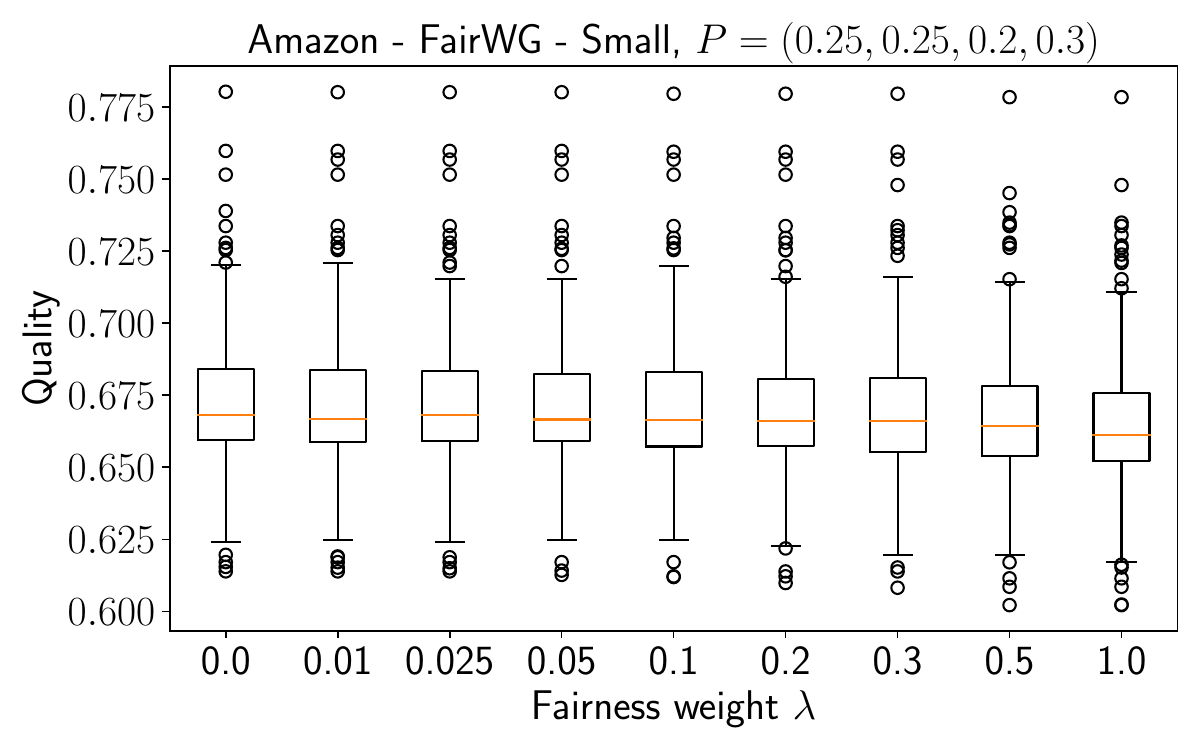}
\end{minipage}
\begin{minipage}{.49\textwidth}
\includegraphics[width=\linewidth]{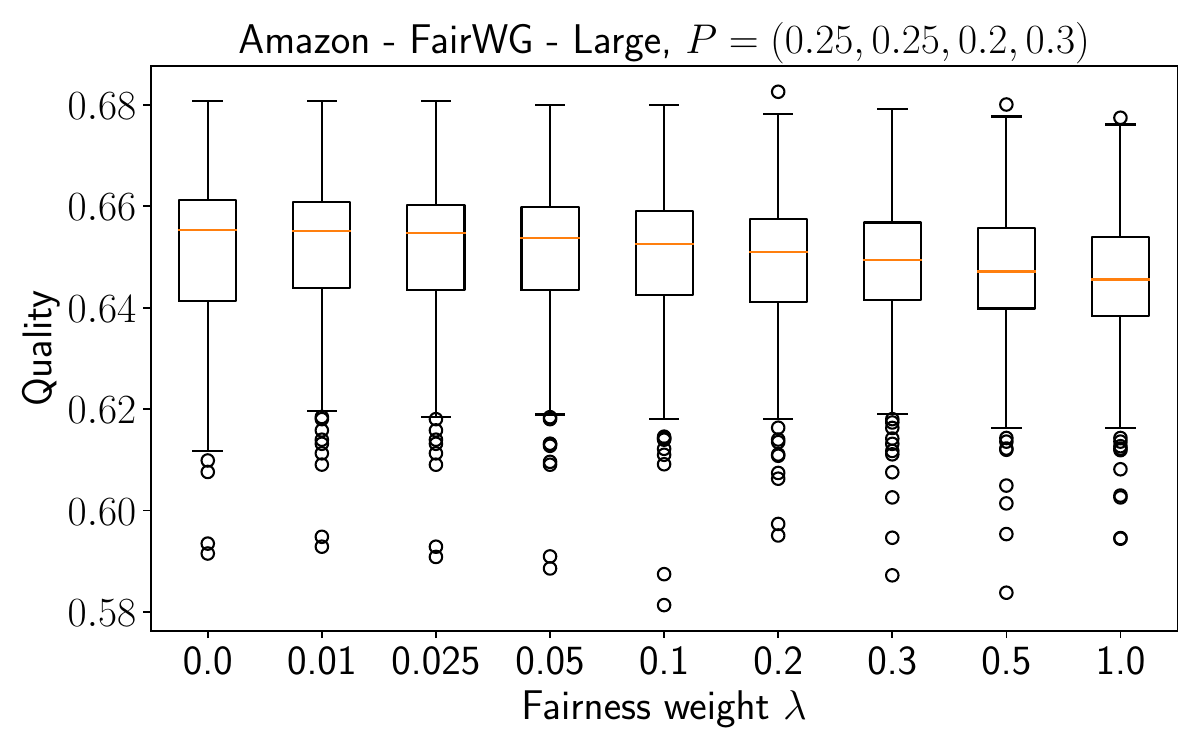}
\end{minipage}
\begin{minipage}{.49\textwidth}
\includegraphics[width=\linewidth]{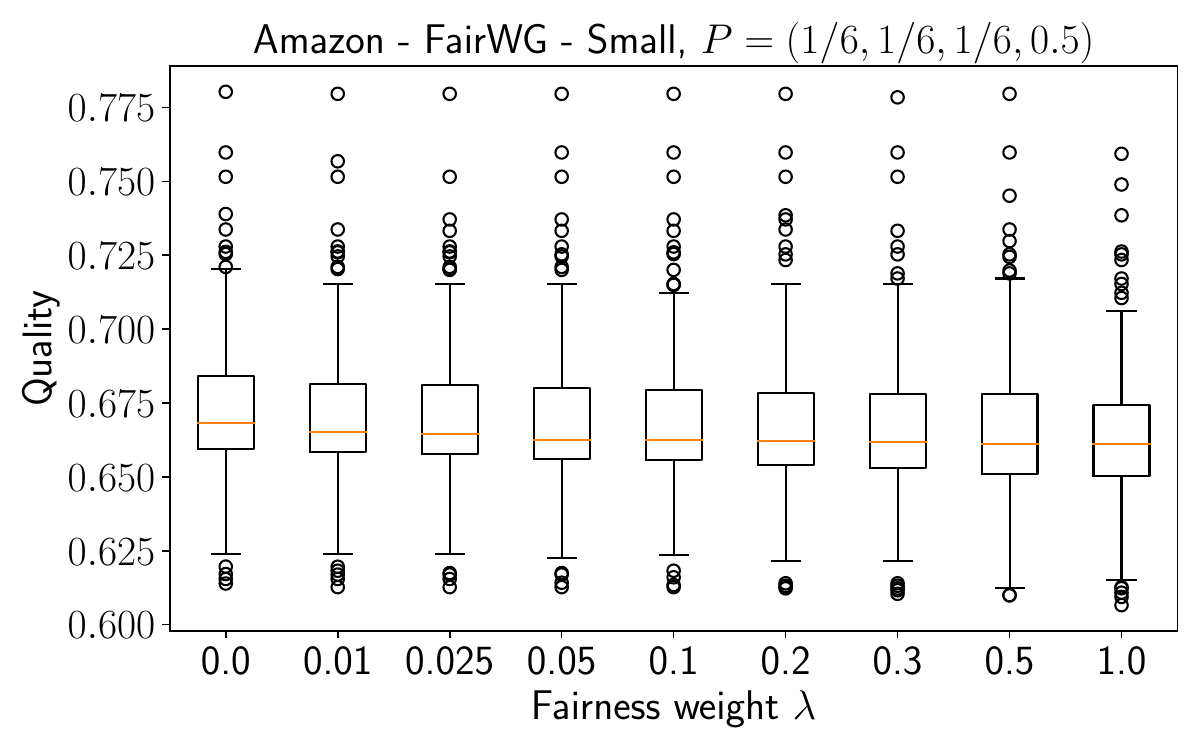}
\end{minipage}
\caption{\textsc{FairWG} - Fairness and quality on \textit{Amazon}.}
\label{fig:fbp_amazon}
\end{figure}

\begin{figure}[tt]
\centering
\includegraphics[width=\linewidth]{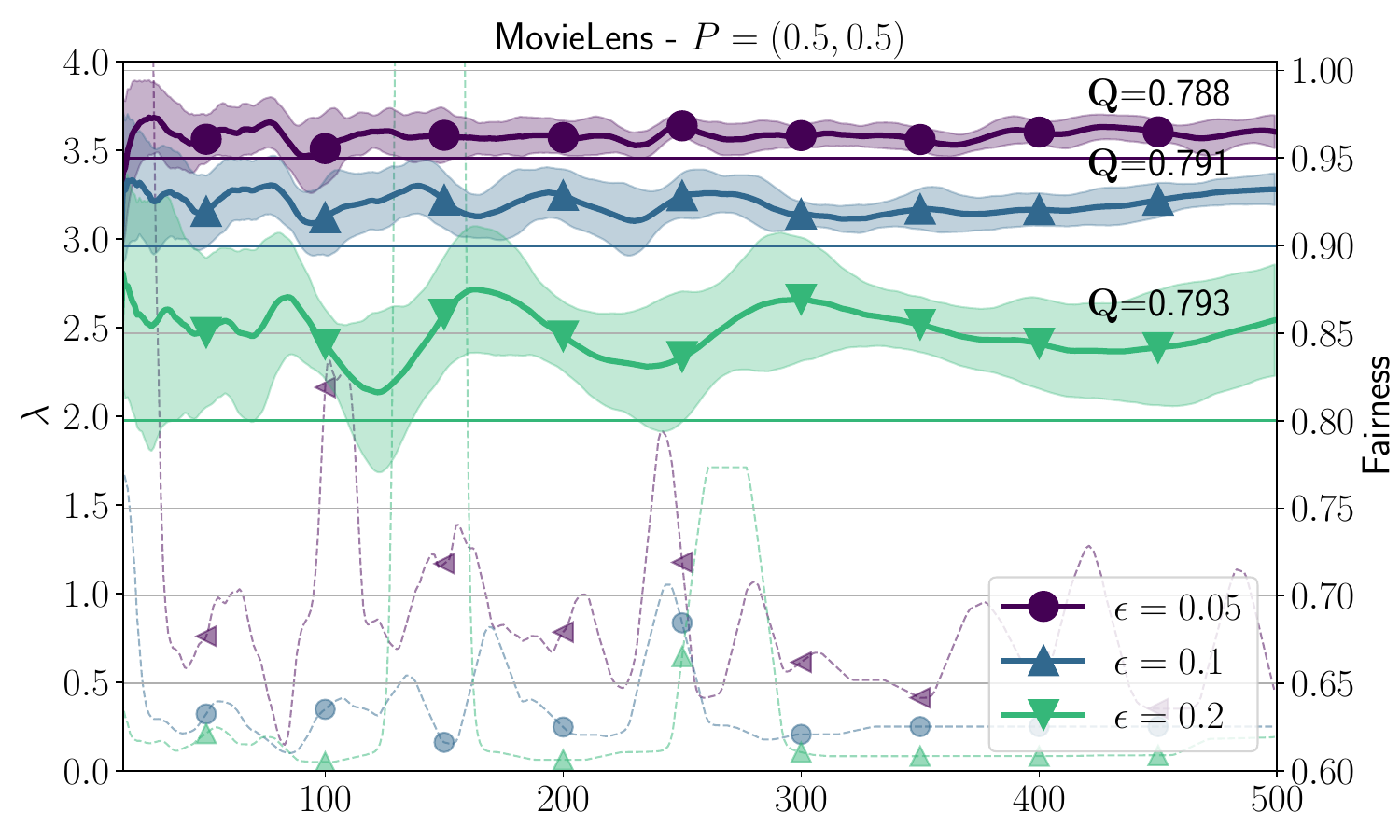}
\caption{Results for \textsc{AdaFairWG} with tolerance levels $\epsilon \in \{5\%, 10\%, 20\%\}$ and horizon $T=500$, for \textit{MovieLens}. Plain lines: fairness $\mathbf{F}$. Dashed lines: adaptive fairness weight $\lambda$. Beside each curve, the average quality $\mathbf{Q}$ is displayed.}
\label{fig:adafairwg_ml}
\end{figure}

\end{document}